\documentclass[preprint,12pt]{elsarticle}

\usepackage{amssymb}
\usepackage{amsmath}

\usepackage{lmodern}  
\usepackage{array}  
\usepackage{makecell}
\usepackage{amsmath}
\usepackage{amsfonts}
\usepackage{amssymb}  
\usepackage{mathrsfs}  
\usepackage{graphicx}
\usepackage{caption}
\usepackage{soul}
\usepackage{booktabs}
\usepackage{multirow}
\usepackage{multicol}  
\usepackage{array, makecell} 
\usepackage{tabularx}  
\usepackage[cal=boondox]{mathalfa}

\usepackage{stackengine}
\stackMath
\newcommand\tsup[2][2]{%
 \def\useanchorwidth{T}%
  \ifnum#1>1%
    \stackon[-.5pt]{\tsup[\numexpr#1-1\relax]{#2}}{\scriptscriptstyle\sim}%
  \else%
    \stackon[.5pt]{#2}{\scriptscriptstyle\sim}%
  \fi%
}

\newcommand{\CuiEtAl}{Cui \textit{et~al.}}
\usepackage[table]{xcolor}
\usepackage{pgf}
\usepackage{collcell}
\usepackage{booktabs}

\newcommand*{\MinNumber}{0}%
\newcommand*{\MaxNumber}{1}%

\newcommand{\ApplyGradient}[1]{%
	\pgfmathsetmacro{\PercentColor}{100.0*(#1-\MinNumber)/(\MaxNumber-\MinNumber)}%
	\edef\x{\noexpand\cellcolor{black!\PercentColor}}\x\textcolor{black}{#1}%
}
\newcolumntype{R}{>{\collectcell\ApplyGradient}{r}<{\endcollectcell}}

\journal{Applied Soft Computing}

\begin{document}

\begin{frontmatter}



\title{Gradient-Based Neuroplastic Adaptation for the Concurrent Optimization of Neuro-Fuzzy Networks}

\author[wku]{John Wesley Hostetter}
\author[ncsu]{Min Chi}
\affiliation[wku]{organization={Department of Analytics and Information Systems at Western Kentucky University},
            addressline={410 Regents Ave},
            city={Bowling Green},
            postcode={42101},
            state={KY},
            country={USA}}

\affiliation[ncsu]{organization={Department of Computer Science at North Carolina State University},
            addressline={909 Capability Drive},
            city={Raleigh},
            postcode={27606},
            state={NC},
            country={USA}}



\begin{abstract}
Neuro-fuzzy networks (NFNs) are transparent, symbolic, and universal function approximations that perform as well as conventional neural architectures, but their knowledge is expressed as linguistic IF-THEN rules. 
Despite these advantages, their systematic design process remains a challenge. 
Existing work will often sequentially build NFNs by inefficiently isolating parametric and structural identification, leading to a premature commitment to brittle and subpar architecture. 
We propose an application-independent approach called \textit{gradient-based neuroplastic adaptation} for the \textit{concurrent optimization} of NFNs' parameters and structure. By recognizing that NFNs' parameters and structure should be optimized simultaneously as they are deeply conjoined, settings previously unapproachable for NFNs are now accessible, such as the online reinforcement learning of NFNs for vision-based tasks. 
The effectiveness of concurrently optimizing NFNs is empirically shown as it is trained by online reinforcement learning to proficiently play challenging scenarios from a vision-based video game called DOOM.
\end{abstract}

\begin{graphicalabstract}
\centering
\includegraphics[scale=0.965]{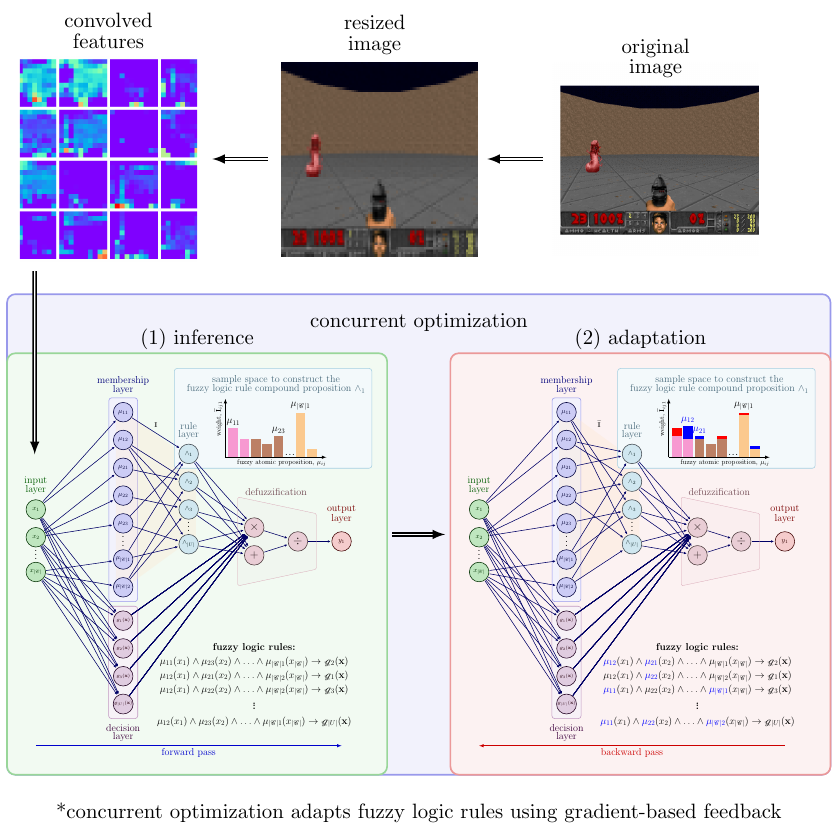}
\end{graphicalabstract}

\begin{highlights}
\item This work concurrently optimizes neuro-fuzzy networks’ parameters and structure.
\item Our \textit{gradient-based neuroplastic adaptation} adjusts fuzzy rules based on performance.
\item Our work integrates with arbitrary operations (e.g., convolutional neural networks).
\item Hierarchical neuro-fuzzy networks are easily constructed via our proposed approach.
\item Entire design of neuro-fuzzy networks with online reinforcement is now trivial.
\end{highlights}


\end{frontmatter}



\section{Introduction}\label{introduction}
Deep neural networks (DNNs) 
often rely upon large quantities of data to effectively train and learn their parameters, which is sometimes impractical in particular domains \cite{ye_mastering_2021}. Instead, some proponents argue that a greater focus should be placed on data-centric methodologies \textemdash{} techniques that can still yield effective and accurate models from limited but higher quality data \cite{mazumder2023dataperf}. Further, a shift to a new research trend has begun, emphasizing developing models that perform well yet remain transparent, interpretable, or even better, explainable to humans, despite the problem's complexity \cite{arrieta_explainable_2019}.

\textit{Neuro-fuzzy networks (NFNs)}, by contrast, are sample-efficient neuro-symbolic architectures proposed to improve transparency and interpretability \cite{aric_2}. The development of these neuro-symbolic architectures has historically relied on pre-existing knowledge, often expressed as readable IF-THEN rules, sourced from human domain experts. The collection of these rules creates a knowledge base written in linguistics to describe an approximate and imprecise causality between stimuli and response; these are called \textit{fuzzy logic rules}, as a special form of logic is employed to handle the imprecise meaning and nature of language. NFNs combine the advantages of DNNs while emphasizing transparent knowledge and have the property of universal function approximation \cite{wang_universal_function_approx, wang_mendel_universal_function_approx, ying_necessary_1994, kosko_1994, zeng_approximation_1995, universal, halgamuge_trainable_1998, alimi_beta_2003}. 

The hurdle to deploying such systems is their inability to readily adapt to an environment, as they were traditionally designed with the intent that a human would meticulously craft their components. 
Today, modern research on NFNs' parametric and structural identification predominantly adopts data-driven approaches, but may still treat these tasks as disjoint: either performing them sequentially or omitting one of the tasks altogether \cite{Azar_2010}. A further dilemma NFNs face is the requirement to address the well-known symbol grounding problem \cite{HARNAD1990335} \textemdash{} a philosophical question regarding how symbols, such as words or abstract representations, acquire meaning in relation to our perceptions of reality. By their very nature, NFNs can fine-tune the relationship between perceptions (i.e., quantitative measurements) and symbols (i.e., qualitative assessments) by adjusting the parameters that underlie their respective mathematical formulas. Still, care must be taken, as NFNs should abide by several constraints to ensure proper control for every possible stimuli. Most notable is $\epsilon$-completeness \cite{lee_flc_12}, where the NFN must have a dominant fuzzy logic rule by requiring each variable to have at least one linguistic term that is activated greater than $\epsilon$ to describe the input. In other words, to ensure reliable performance, NFNs should aim to have at least one symbol that sufficiently describes each perception that is likely to occur. A potential approach is to generate new symbols to fulfill this need when the existing ones are woefully insufficient. However, NFNs, as a neural architecture, cannot add or remove symbols without assistance from some form of automated mechanism. 
Furthermore, adding or removing symbols introduces a subsequent challenge: determining how to appropriately update the fuzzy logic rules, particularly when incorporating a newly created symbol (i.e., fuzzy atomic proposition).

This paper contains a few unique proposals to the previously mentioned challenges of NFNs, which are inspired by scientific theories regarding the inner workings of the human brain. The most significant contribution is the proposal of an application-independent approach that facilitates the \textit{concurrent optimization} of NFNs' parametric and structural identification, called \textit{gradient-based neuroplastic adaptation (GBNA)}. This method adapts NFNs to different environments by blending related principles from the neuroplasticity of the human brain and incorporating them: \textit{synaptic plasticity} \cite{Citri2008} and \textit{structural plasticity} \cite{Maguire2000-az, Scholz2009, May2011-fm, Zatorre2012, Sampaio-Baptista2017-gr}. For clarity, synaptic plasticity may involve long-term potentiation (LTP) \cite{Bliss1973-ia} or long-term depression (LTD) \cite{MALENKA20045} 
to strengthen connections that facilitate successful learning (via LTP) and weaken those that do not (via LTD). 
Similarly, structural plasticity refers to biological mechanisms regarding \textit{synaptogenesis} (the creation of new synapses) or \textit{neurogenesis} (the creation of new neurons). 
Building on this biological inspiration, the primary contribution of this paper \textemdash{} GBNA \textemdash{} incorporates several key computational strategies designed to emulate neuroplasticity, akin to neural architecture search \cite{zoph2017neuralarchitecturesearchreinforcement}. 


\paragraph{First Contribution} Importantly, we propose treating the discovery of fuzzy logic rules as a sampling problem. This is accomplished by relaxing NFNs' sparse, binary connections that fundamentally represent the fuzzy logic rules to real-valued parameters and interpreting them as weights or probabilities. This transformation addresses the non-differentiability of NFNs' binary connections and then reconstructs them by differentiably sampling the fuzzy logic rules via a straight-through Gumbel estimator (STGE) \cite{jang2017categorical, maddison2017the}. Alternatively, we also consider whether stochastic sampling is even necessary, by exploring the more straightforward straight-through estimator (STE) \cite{bengio2013estimatingpropagatinggradientsstochastic} technique as well. Therefore, either option is considered an implementation of GBNA. If STE is used, the process closely mimics synaptic plasticity as connections that promote successful learning are strengthened (i.e., LTP), eventually discovering the ideal fuzzy logic rules. On the contrary, if STGE governs GBNA, then various compound conditions for fuzzy logic rules are differentiably and stochastically sampled from a probabilistic distribution; STGE may culminate an increase in the exploration of potential fuzzy logic rules but may risk leading to excessive internal thrashing of the NFN's structure or exploring options which negatively affect requirements (e.g., $\epsilon$-completeness). 
Consequently, the NFN's structure (e.g., fuzzy logic rules) can adapt depending on how well the neuro-symbolic architecture performs with standard gradient-based optimization techniques (e.g., Adam \cite{adam_optimizer}). 
These gradients dictate how to update the synaptic weights, or probabilities, based on the NFN's performance for an objective, subsequently reconfiguring the NFN's structure if necessary. The evolution of which fuzzy logic rules are obtained resembles synaptogenesis (or, conversely, synaptic pruning) in the NFN as the physical structure of the neuro-symbolic system adapts to the given task or environment, reflecting changes in its connectivity and function. The overall effects of synaptogenesis are evident during the forward pass of the NFN. At the same time, the dynamics of LTP and LTD manifest in the backward pass, where synaptic weights are adjusted accordingly. 

\paragraph{Second Contribution} The other aspect of structure discovery in NFNs concerns the symbol grounding problem; when the NFN encounters an unrecognized percept (i.e., fails $\epsilon$-completeness), our method emulates neurogenesis \cite{Eriksson1998-aa, Spalding2013-fp, Kempermann2015-cf, Boldrini2018-pa, Sorrells2018-le, Moreno-Jimenez2019-mh} by using an adapted form of Welford's online algorithm to safeguard the NFN to this novel stimuli. Although, as a neuro-symbolic system, the NFN does not necessarily create new \textit{neurons} when existing representations are insufficient. Instead, it generates new ``vague'' \textit{symbols}, accompanied by axonal sprouting \cite{RAISMAN196925, KAAS2000173}, to form exploratory connections with the fuzzy logic rules in which these symbols may serve as premises. We empirically show that a \textit{delayed} neurogenesis to satisfy NFN's need for $\epsilon$-completeness may be sufficient or warranted when concurrently optimizing NFNs. Delaying neurogenesis, rather than greedily acting upon it, ensures newly added ``vague'' symbols (i.e., fuzzy atomic propositions) are sufficiently explored by GBNA and subsequently prevents membership thrashing \textemdash{} where the NFN catastrophically succumbs to the indecisiveness of its internal knowledge base. 

\paragraph{Third Contribution} This paper tackles challenging high-dimensional vision-based tasks to demonstrate the effectiveness of our NFNs. NFNs, or fuzzy inference systems in general, notoriously struggle with high-dimensional inputs as they easily suffer from issues related to numerical underflow \cite{cui_curse_2021}. We investigate how existing techniques in the literature might help in alleviating such concerns in vision-based tasks, but also propose our own solutions based on \CuiEtAl's \cite{cui_curse_2021} observations. Notably, \CuiEtAl~reveal that numerical underflow of Takagi-Sugeno-Kang (TSK) NFNs occurs due to the presence of \texttt{softmax} in determining fuzzy logic rule applicability. However, their attention was primarily focused on the input dimensionality. We offer additional insights that TSK NFNs suffer not only from input dimensionality but also when the number of fuzzy logic rules is significantly high. To remedy this issue, we propose a reasonable alteration to incorporate \textit{layer normalization} \cite{ba2016layer} within TSK NFNs and explore its implications in our experimentation.

\paragraph{Fourth Contribution} We identify and explore an opportunity to enhance the clarity and interpretability of NFN's decision-making. Interestingly, \texttt{softmax} belongs to a family of transforms called $\alpha$\texttt{-entmax} \cite{peters-etal-2019-sparse}. This family of transforms offers a generalization from \texttt{softmax} (when $\alpha$=1) to \texttt{sparsemax} \cite{10.5555/3045390.3045561} (when $\alpha$=2). The significance of this realization is that \texttt{softmax} will not assign a zero value to an entry despite insufficient support. In other words, standard TSK NFNs ($\alpha$=1) allow fuzzy logic rules to have activations that are \textit{extremely} near zero, but still non-zero. This is distracting for the purpose of interpretability, as it becomes impossible to entirely dismiss a given rule's effect on the overall TSK NFN's output, albeit, even if its influence was rather dismal. However, if we replaced \texttt{softmax} with $\alpha$\texttt{-entmax} and encouraged sparsity ($\alpha > 1$), then the TSK NFN would naturally truncate a fuzzy logic rule's activation to zero if its support is negligible; thus, ultimately enhancing NFN interpretability by encouraging sparsity in firing levels. For instance, a NFN might have a rather large number of fuzzy logic rules (e.g., 2056), but only a small handful (e.g., 3) would be activated at any given time instead of all of them being activated to some level of degree. 

Overall, the culmination of our first and second contributions concurrently optimizes the NFNs' structure and parameters. The third contribution address computational challenges in high-dimensional fuzzy inference, whereas the fourth contribution further enhances the interpretability of our NFN's decision-making if it is possible to utilize $\alpha$\texttt{-entmax} with $\alpha > 1$. 
This paper introduces potentially the first NFN that can seamlessly integrate with complex or arbitrary neural architectures, such as convolutional neural networks (CNNs), without requiring a predefined structure or relying on ad hoc strategies. The framework addresses critical limitations in current approaches by uniquely enabling the full training of an NFN’s structure and parameters for vision-based tasks according to any learning paradigm. This is achieved by introducing mechanisms to add the necessary but previously missing symbols for constructing fuzzy logic rules and enabling the NFN to reconfigure itself dynamically based on gradient information. Thus, we propose starting with a preliminary random NFN, adding new symbols (i.e., words/linguistic terms) as needed, and modifying the rules' conditions based on the NFN's performance on a given objective function. The effectiveness of concurrent optimization is demonstrated by training hierarchical NFNs to play various scenarios in a vision-based video game, DOOM, using an online reinforcement learning (RL) algorithm.

\section{Background \& Related Work}\label{section:related_work}

\subsection{Fuzzy Inference Systems}
Intelligent systems have greater 
\textit{approximate reasoning} by modeling reality with fuzzy logic \cite{is_there_a_need_for_fuzzy_logic}. 
For instance, neural networks with certain characteristics
are provably fuzzy systems; that is, they are mathematically equivalent to a system that uses fuzzy logic \cite{1209317,fls_ann_equivalence}. 
A fuzzy inference system (FIS) is a type of universal function approximation that allows the designer to incorporate any explicit, a priori knowledge into the agent's behavior by using linguistic rules; these rules are traditionally written by human experts in the form of IF-THEN statements \cite{zadeh_book}. As expected, a FIS extends traditional decision tables by inferring its decisions on partially true/applicable IF-THEN statements with fuzzy logic.  
The literature is extensive regarding FISs, with different types of fuzzy logic rules being offered, as well as numerous ways of defining relevant fuzzy logic operations.

\subsection{Neuro-Fuzzy Networks}
A clear advantage of DNNs to FISs is their computational efficiency. Naïve implementation of FISs can become computationally expensive very quickly and may even require the incorporation of specialized hardware \cite{fis_systolic_hardware,KELLER19921,keller_backpropagation_1992}. Additionally, DNNs have also enjoyed widespread success and adaptation thanks to the relative ease of applying them to learn complex tasks when given sufficient data and network complexity (e.g., number of hidden layers/neurons, reasonable selection of their activation functions). On the contrary, the structure as well as parameters of a FIS are often historically designed based on human intelligence \cite{lee_flc_12,jammu_fpga_2016}. Therefore, it would be logical and natural to want to extend FISs with this same learning ability to remove the dependency on meticulously designing their components. 

Neuro-fuzzy networks (NFNs) are computationally efficient, connectionist representations of FISs that allow gradient descent to fine-tune parameters such as those that define fuzzy atomic propositions used in fuzzy logic rules' conditions or decisions \cite{lin_real-time_1992,jang_anfis_1993}. NFNs merge symbolic inference with neural reasoning and allow explicit as well as implicit knowledge to be utilized. In contrast, DNNs typically can only use implicit knowledge extrapolated from large amounts of data. On a high level, an NFN is a sparsely connected neural network where weights between the layers are restricted to binary values. Non-standard activation functions are used throughout, where each layer directly corresponds to a step of the FIS; details are included in Section~\ref{appendix:nfn}. 

Their connection to neural networks enabled more efficient parameter tuning by adjusting fuzzy atomic propositions based on the NFN's performance relative to a given objective function. By representing a FIS in this manner, the parameters (and later the structure, as proposed in this paper) of the FIS may be fine-tuned with gradient descent. Thus, an objective of this connectionist representation is to be efficient and remain differentiable. Most NFN implementations can only achieve differentiability of the parameters, further highlighting the novelty and potential impact of this paper, as our proposed changes allow concurrent optimization of parameters and structure. 
Overall, NFNs differ from conventional DNNs as they allow both explicit and implicit knowledge to be utilized. In contrast, DNNs typically can only use implicit knowledge extrapolated from large amounts of data.

The effectiveness of a FIS or NFN largely depends on its design; namely, for an agent to influence or control its environment adequately, the number of fuzzy atomic propositions for each state input must be determined, as well as their corresponding membership functions to define the fuzzy atomic propositions themselves \cite{zadeh_book}. Additionally, the membership functions can take on various shapes, are typically problem-dependent, and may impact computational cost \cite{klir_yuan}. The choice for the number of fuzzy atomic propositions then influences the number of fuzzy logic rules required to complete the FIS, and the size of the agent's knowledge base may grow exponentially with respect to the fuzzy atomic propositions defined, as well as the dimensionality of the input space domain \cite{zhou_popfnn_1996}. Therefore, it is critical to identify the minimum number of fuzzy atomic propositions or fuzzy rules to control the environment sufficiently to maintain linguistic interpretability. 

Fuzzy modeling research of FISs or NFNs has a rich literature on constructing fuzzy systems through various approaches:
\begin{itemize}
    \item \textit{Expert-design} involves manually constructing FISs or NFNs by relying on a human expert to explicitly design every component. This method, which characterized the original conception of FISs, has become less practiced as the field has modernized and shifted toward more data-driven methodologies. 
    
    \item \textit{Distillation \& Extraction} often involves observing DNNs' input-output behavior and then extracting fuzzy logic rules accordingly \cite{extract_fuzzy_rules, hostetter2023leveraging}. However, such methods usually only result in an approximate representation of the original model \cite{rules_dnn}.

    \item \textit{Transformation via Mathematical Equality} may yield a FIS or NFN from another model, often by exploiting explicit mathematical properties. However, there are significant limitations to this approach. The transformation's applicability often comes with several caveats, the extracted knowledge may remain obfuscated due to the applied transformation, exponential growth of fuzzy logic rules may occur, or non-standard fuzzy logic conventions may be employed \cite{are_ann_black_boxes, are_ann_white_boxes}; thus, these findings remain more theoretically insightful than practical.
    
    \item \textit{Evolutionary} approaches are often explored as a possibly viable strategy in designing FISs or NFNs. A vast body of research has explored the use of genetic algorithms for designing and optimizing fuzzy systems; for instance, genetic fuzzy systems have successfully reduced fuzzy logic rule complexity and performed well in supervised learning settings involving high-dimensional regression datasets \cite{adeli_machine_1994, mitra_neuro-fuzzy_2000, tung_gensofnn_2002, cordon_ten_2004, kubota_genetic_2005, melin_analysis_2007, al-shamri_fuzzy-genetic_2008, khayat_novel_2009, chua_new_2009, shill_optimization_2013, dahal_ga-based_2015, ohashi_learning_2019,goharimanesh_fuzzy_2020, navarro-almanza_interpretable_2022, park_generative_2023}. However, these methodologies often rely on a slightly more informed trial-and-error process, involving extensive steps to compute fitness functions and requiring meticulous crossover manipulations. Approaches like MOKBL+MOMs \cite{aghaeipoor_mokblmoms_2019} require complex, multi-step procedures to identify and prune fuzzy logic rules with appropriate fitness function definitions; moreover, a single optimization run can take several hours, as numerous candidates must be evaluated to refine the FIS or NFN \cite{aghaeipoor_mokblmoms_2019}. Genetic algorithms' computationally intensive nature makes them impractical for our use case, which involves applying NFNs to complex, reward-based scenarios that depend on RL.
    
    \item \textit{Self-Organization} is a compelling alternative to evolutionary approaches, where the FIS or NFN are incrementally and/or iteratively built with algorithms dedicated to produce components, such as fuzzy atomic propositions (e.g., modified Kohenen's feature-maps algorithm \cite{lin_neural-network-based_1991}, unsupervised discrete clustering technique \cite{singh_dct-yager_2008}) or fuzzy logic rules (e.g., competitive learning law \cite{lin_neural-network-based_1991}). The culmination of these algorithms' outputs then result in the designed fuzzy system. These techniques to construct an NFN may apply to unsupervised \cite{sevakula_fuzzy_2015,hostetter2023latent,bolat_interpreting_2020,zhao_self-organized_2022}, supervised \cite{lin_real-time_1992,jang_anfis_1993}, or reinforcement learning \cite{lin_reinforcement_1994,rfalcon,self_organizing_fql,hostetter2023self} paradigms. Unfortunately, existing literature is often tailored to a specific learning strategy to construct the NFN more easily. For instance, most existing research into self-organizing NFNs is restricted to supervised learning \cite{kasabov_denfis_2002,wang_self-adaptive_2002,allende-cid_self-organizing_2008,tung_safin_2011} as it assumes labeled data availability and uses those labels to guide the construction (e.g., self-organizing maps \cite{zia_neuro-fuzzy_1994}, rule simplification \cite{ang_rspop_2005}, supervised discrete clustering technique \cite{singh_dct-yager_2008}). Pre-existing strategies for the gradient-based adjustment of fuzzy logic rules in NFNs are limited by their impractical assumptions or inability to sufficiently scale well; notably, \textit{most} works (e.g., POPFNN family \cite{zhou_popfnn_1996,zhou_pseudo_1996,quek_popfnn-aars_1999,ang_popfnn-cris_2003}) must consider \textit{all} possible fuzzy logic rules \cite{zhou_popfnn_1996, ang_popfnn-cris_2003, singh_dct-yager_2008}, which grow exponentially with respect to input dimensionality and fuzzy atomic proposition count. However, with LazyPOP \cite{zhou_pseudo_1996}, the number of fuzzy logic rules considered then scales with respect to the size of training data. More recent research (e.g., RSPOP-CRI \cite{ang_rspop_2005}) incorporates strategies from rough set theory \cite{pawlak1991roughsets} to reduce the number of fuzzy logic rules by leveraging a KR-system to facilitate rule identification, attribute reduction, and rule reduction \cite{ang_rspop_2005}. RSPOP-CRI's rule identification is arguably superior to LazyPOP since it does not require user-defined criteria. Published state-of-the-art research still relies on RSPOP variations, such as ARPOP \cite{cheu_arpop_2012}, ieRSPOP \cite{das_ierspop_2016}, and PIE-RSPOP \cite{iyer_pie-rspop_2018}. However, RSPOP and its variations rely on labeled data, so it is inappropriate for other learning paradigms (e.g., RL). 
\end{itemize}

\subsection{Fuzzy Reinforcement Learning}
Fuzzy RL was introduced to incorporate existing linguistic control knowledge to reduce the time required to solve RL, buts its resulting policies are also interpretable \cite{aric_2}. GARIC could learn despite weak reinforcements, but was online-only as its on-policy actor-critic framework was dependent on exploration \cite{berenji_learning_1992}. 
Fuzzy Q-Learning \cite{fql_and_dynamic_fql} extends Q-Learning \cite{Watkins1992} with fuzzy logic and is an online, off-policy method for tasks with continuous state space and either continuous or discrete actions \cite{glorennec_fuzzy_1997}. An important stipulation of these works is the assumptions that experts can manually provide fuzzy sets and fuzzy logic rules. Self-organizing NFNs by Q-learning typically build the fuzzy logic rules and learns their corresponding Q-values via \emph{online} interaction by trial and error \cite{kasabov_denfis_2002}. 
Fuzzy particle swarm RL self-organizes NFNs with model-based offline RL but requires a simulation to learn its parameters and must define sought-after fuzzy logic rule count \cite{hein_particle_2017}. 
``Interpretable fuzzy reinforcement learning (IFRL)'' was recently coined \cite{huang_interpretable_2020} where empirical fuzzy sets \cite{angelov_empirical_2018} construct NFNs for a model-free actor-critic. However, IFRL is only demonstrated in the online setting, and once again, NFNs must be constructed \textit{beforehand} and cannot be applied to computer vision settings.

\section{Neuro-Fuzzy Networks}\label{appendix:nfn}
Let $U$ uniquely identify each fuzzy logic rule, and $\mathcal{C}$ as well as $\mathcal{D}$ reference condition and decision attributes, respectively, such that $\mathcal{A} = \mathcal{C} \cup \mathcal{D}$. 

This paper only considers TSK FISs with Gaussian fuzzy sets, singleton fuzzification and product inference engine as several powerful computational tricks can then be leveraged for high-dimensional fuzzy inference. 
However, it is important to note that GBNA would work for Mamdani FISs, as well. We define our TSK FISs as

\begin{equation}
\label{eq:tsk_fuzzy_system:coa_defuzzifier}
    \mathfrak{F}_{\text{TSK}}(\mathbf{x}) = \frac{ 
    \sum\limits_{u \in U} \mathcal{g}_{u}(\mathbf{x})     
    \big (
    \prod\limits_{i \in I_{\mathcal{C}}} 
    \mathcal{m}_{u}(i)(x_{i}) 
    \big )
    }{ 
    \sum\limits_{u \in U}
    \big (
    \prod\limits_{i \in I_{\mathcal{C}}} 
    \mathcal{m}_{u}(i)(x_{i}) 
    \big )
    }
\end{equation}
\noindent
such that the relevance of each fuzzy logic rule $u \in U$ is determined by 
$\prod_{i \in I_{\mathcal{C}}} 
\mathcal{m}_{u}(i)(x_{i})$, and each $\mathcal{m}_{u}(i)$ retrieves a Gaussian fuzzy set, $\mu_{ij}$:\begin{equation}\label{eq:mf:gaussian}
	\mu_{\textit{gaussian}}(x, c, \sigma) = \exp \Bigg (- \frac{(x - c)^{2}}{2{\sigma}^{2}} \Bigg)
\end{equation}
\text{where $c$ represents the center and $\sigma$ is the width.} 
For convenience, let $\mu_{iju} = \mathcal{m}_{u}(i)$ such that $c_{iju}$ and $\sigma_{iju}$ are its center and width, respectively. We may then simply rewrite Eq.~\ref{eq:tsk_fuzzy_system:coa_defuzzifier} for complete clarity to yield 
\begin{equation}
\label{eq:tsk_fuzzy_system:coa_defuzzifier_expanded}
    \mathfrak{F}_{\text{TSK}}(\mathbf{x}) = \frac{ 
    \sum\limits_{u \in U} \mathcal{g}_{u}(\mathbf{x})     
    \Big (
    \prod\limits_{i \in I_{\mathcal{C}}} 
    \exp \big (-\frac{({x}_{i} - {c}_{iju})^2}{2\sigma_{iju}^2} \big )
    \Big )
    }{ 
    \sum\limits_{u \in U}
    \Big (
    \prod\limits_{i \in I_{\mathcal{C}}} 
    \exp \big (-\frac{({x}_{i} - {c}_{iju})^2}{2\sigma_{iju}^2} \big )
    \Big )
    }
\end{equation}
which will become beneficial later as we begin to utilize particular properties for high-dimensional fuzzy inference. Each step of the fuzzy inference can be represented by a neural architecture \textemdash{} an NFN. An NFN typically has the following layers: (1) Fuzzification Layer (not required here), (2) Condition Layer, (3) Rule Layer, (4) Decision Layer, and (5) Defuzzification. These layers correspond directly to an individual step in FIS inference. For notational simplicity, we will assume a batch size of 1 in the following mathematics.

\subsection{Describing the Input with Linguistic Terms}
\paragraph{Condition Layer} Contains \textit{all} the linguistic terms that describe the input space (i.e., $\bigcup_{i \in I_{\mathcal{C}}} \mathcal{M}_{i}$). Edges from the input layer to the condition layer \textit{only} exist with a value of $1$ if and only if the linguistic term applies to that linguistic variable. To clarify, $x_{i}$ for $i \in I_{\mathcal{C}}$ would only be connected to each $\mu_{ij}$ from its associated fuzzy sets, $\mathcal{M}_{i}$. 
This layer's output is the collection of all memberships to all linguistic terms across the linguistic variables for condition attributes $\mathcal{C}$, and may be represented as a matrix:
\begin{equation}
\label{eq:nfn:membership_matrix}
\boldsymbol{\mu}_{\mathcal{C}}(\mathbf{x})
= \bigl[\mu_{ij}(x_i)\bigr]_{
    \substack{
        i = 1,\dots,|\mathcal{C}| \\
        j = 1,\dots,|\mathcal{M}_i|
    }}
\end{equation}
where $|\mathcal{C}|$ is the count of condition attributes. In the event some $\mu_{ij}({x}_{i})$ does not exist for the index pair $(i, j)$, then a special flag (e.g., \texttt{NaN}, -1) may be placed to occupy the location in the meantime. 
An accompanying \textit{membership mask}, $\mathbf{M}$, of shape $|\mathcal{C}| \times \max_{i \in I_{\mathcal{C}}}(|\mathcal{M}_{i}|)$ may exist if necessary which contains a $1$ if and only if a linguistic term exists at entry $i, j$, else, it is $0$. Non-existing memberships may cautiously be dropped by Hadamard product (element-wise multiplication, $\odot$) 
such that $\boldsymbol{\mu}_{\mathcal{C}}(\mathbf{x}) \odot \mathbf{M}$. 

\subsection{Activating the Fuzzy Logic Rules}
\paragraph{Rule Layer} Hosts the fuzzy logic rules' compound conditions; the inputs provided to it are from Eq.~\ref{eq:nfn:membership_matrix}. 
Similar to the condition layer, they operate only on their respective linguistic variables and terms involved in the compound conditions. As a result, edges are assigned a value of $1$ between all relevant (fuzzy) condition propositions, while all other connections are set to $0$. This paper only focuses on NFNs that utilize a singleton fuzzification, product inference engine, and Center of Area defuzzification. Thus, each node $u$ of this layer calculates $\prod_{i \in I_{\mathcal{C}}} \mathcal{m}_{u}(i)(x_{i})$ where $\mathcal{m}_{u}(i)$ retrieves the appropriate linguistic term for the $i$\textsuperscript{th} attribute used by fuzzy logic rule $u$. The nodes' outputs correspond to the activation of each fuzzy logic rule in the FRB regarding the compound condition attributes applicability to input $\mathbf{x}$. 

\subsection{Determining the Decisions}
\paragraph{Decision Layer} In this work, we employ TSK-type NFNs, where the recommended decision of the $u$\textsuperscript{th} fuzzy rule is $\mathcal{g}_{u}(\mathbf{x}) = \mathbf{W}_{u}\mathbf{x} + \mathbf{b}_{u}$.

\subsection{Calculating the Output}
\paragraph{Defuzzification} 
Finally, the rest of Eq.~\ref{eq:tsk_fuzzy_system:coa_defuzzifier} is then carried out (e.g., weigh each fuzzy logic rule decision by its applicability, sum over the rules, normalize). 
\subsection{An Important Rewrite of Takagi-Sugeno-Kang Neuro-Fuzzy Networks}\label{appendix:computational_tricks}
\CuiEtAl~reveals the presence of \texttt{softmax} in TSK FISs by rewriting the formula as $\mathfrak{F}_{\text{TSK}}(\mathbf{x}) = \sum_{u \in U} \mathcal{g}_{u}(\mathbf{x}) \overline{w}_{u}(\mathbf{x})$ such that


\begin{minipage}{0.475\linewidth}
    \begin{equation}\label{eq:softmax}
    \overline{w}_{u}(\mathbf{x}) = \frac{\exp \big (
    {w}_{u}(\mathbf{x})
    \big)
    }{\sum\limits_{u \in U} \exp \big({w}_{u}(\mathbf{x})\big)}
    \end{equation}
\end{minipage}\hfill
\begin{minipage}{0.475\linewidth}

    \begin{equation}
    \label{eq:mu_rule_sum}
    w_{u}(\mathbf{x}) = - \sum\limits_{i \in I_{\mathcal{C}}} 
    \frac{({x}_{i} - {c}_{i, j, u})^2}{2\sigma_{i, j, u}^2}
\end{equation}
\end{minipage}
\noindent
and $w_{u}(\mathbf{x})$ is the (preliminary) activation of fuzzy logic rule $\mathcal{f}_{u}$. 
Thus, Eq.~\ref{eq:softmax} is \texttt{softmax} and numeric underflow can be avoided with a common trick \cite{cui_curse_2021}

\begin{equation}\label{eq:norm_softmax}
\overline{w}_{u}(\mathbf{x}) = \frac{\exp \big (
{w}_{u}(\mathbf{x}) - \max{w}_{u}(\mathbf{x})
\big)
}{\sum_{u}^{U} \exp \big({w}_{u}(\mathbf{x}) - \max{w}_{u}(\mathbf{x}) \big)}
\end{equation}
where $\max {w}_{u} = \max_{u \in U} {w}_{u}$. The TSK NFN's operations are updated to utilize this rewrite accordingly without altering the overall fuzzy inference.

\section{Methodology}\label{method}
We developed concurrent optimization of NFNs to make them more flexible and compatible with complex stimuli, such as observed images in vision-based tasks. We also sought an approach that allowed arbitrary learning techniques to train the NFN, such as online/offline RL, imitation learning, and supervised learning. Pursuing these objectives, we explored and thoroughly evaluated numerous approaches while developing the concurrent optimization of NFNs. This section provides a comprehensive account of the design decisions we investigated and detailed rationales. 


\subsection{Gradient-Based Neuroplastic Adaptation}
\label{section:gumbel}
Gradient-based neuroplastic adaptation (GBNA) can be implemented either with STE or STGE. The former leverages strong connections that yield successful fuzzy logic rules, whereas the latter interprets the real-valued weights as probabilities. In STGE, fuzzy logic rules are differentiably and stochastically sampled from these probabilistic distributions, allowing for greater exploration. In contrast, STE behaves in a more greedy manner. 

\subsubsection{Relaxing The Architecture} 
Let $\mathcal{C}$ and $U$ denote the sets for condition (input) attributes and unique universal identifiers for fuzzy logic rules, respectively. The index set of $\mathcal{C}$ will be denoted as $I_{\mathcal{C}}$. Each $i^{\text{th}}$ condition attribute, $\mathcal{a}_{i}$, for $i \in I_{\mathcal{C}}$, has a set of symbols, $\mathcal{M}_{i}$ that linguistically describe $\mathcal{a}_{i}$. For instance, $\mathcal{a}_{i}$ could be the attribute ``temperature'', and $\mathcal{M}_{i} = \{ \text{cold, hot} \}$, such that each symbol in $\mathcal{M}_{i}$ is a fuzzy atomic proposition. A matrix, $\mathbf{I}$, may be constructed representing connections between the condition layer (i.e., input linguistic terms) and their associated fuzzy logic rules for which they are involved. 
Each entry at $\mathbf{I}_{i, j, u}$ is $1$ if and only if the ${j}^{\text{th}}$ linguistic term of the ${i}^{\text{th}}$ attribute is a premise/condition of the ${u}^{\text{th}}$ rule,  otherwise it is $0$. 

NFNs' binary weight matrices, particularly $\mathbf{I}$, make it difficult to rearrange fuzzy logic rules' conditions in response to their performance. We propose to \textit{relax} these constraints temporarily, such that $\mathbf{I}$ is substituted with $\tsup[1]{\mathbf{I}}$, where all entries in $\tsup[1]{\mathbf{I}}$ may be assigned any real-number; these values may be interpreted as weights with STE, or in STGE, entries of $\tsup[1]{\mathbf{I}}$ may represent \textit{logits}, or raw non-normalized probabilities that the edge exists; this work uses the Xavier normal distribution to initialize $\tsup[1]{\mathbf{I}}$. Thus, with STE, $\mathbf{I}$ is differentiably reconstructed only from the \texttt{argmax} of $\tsup[1]{\mathbf{I}}$'s strongest weights. On the contrary, STGE will differentiably sample different possible conditions that should be explored by relying on $\tsup[1]{\mathbf{I}}$ as a matrix of logits; this sampling may also exhibit stochastic behavior when introducing Gumbel noise. The presumed benefit of STGE over STE is that it may allow for greater exploration of alternative combinations for conditions of fuzzy logic rules. However, it could also be possible for the Gumbel noise, and consequently, the stochastic nature during the forward pass, to interfere with stability or convergence. Therefore, we explored both STE and STGE as options for GBNA.

\subsubsection{Incorporating Differentiable Stochastic Sampling} 
The STGE aimed to ensure that the sampling process from a categorical distribution remains differentiable \cite{gumbel_max_trick}. Within the context of NFNs, this amounts to sampling compound condition attributes for fuzzy logic rules in a differentiable way, where each condition attribute has a categorical distribution containing the possible linguistic terms. This allows the gradient-based feedback regarding the NFN's performance to guide which fuzzy logic rules are sampled. Eventually, this sampling converges to the NFN's final set of fuzzy logic rules. 
Noise $\mathbf{N}$ is sampled from a Gumbel distribution (location=0, scale=1.0) when training the NFN. 
The sampled noise $\mathbf{N}$ is added for stochasticity, and the distribution is softened with a temperature parameter, $\tau > \mathbb{R}^{+}$, according to 
\begin{equation}
    \label{eq:add_noise_soften_distribution}
    \tilde{\mathbf{I}}' = \frac{\tsup[1]{\mathbf{I}} + \mathbf{N}}{\tau^2}.
\end{equation}
When evaluating or no longer training the NFN, $\mathbf{N}$ is $\mathbf{0}$. 
Since Gumbel-\texttt{softmax} may yield non-zero probability for invalid selections as it approximates a categorical distribution, 
these may be pushed to 0 with $\mathbf{M'}$ as a binary constraint matrix to filter out nonexistent fuzzy sets (see~\ref{appendix:constraint}):
\begin{equation}
    \label{eq:restricted_gumbel_dist}
    \varphi (\tilde{\mathbf{I}}') = \frac{\exp(\tilde{\mathbf{I}}') \odot \mathbf{M'}}{\sum_{j=1}^{\max_{i \in I_{\mathcal{C}}}(|\mathcal{M}_{i}|)} \exp(\tilde{\mathbf{I}}') \odot \mathbf{M'}}
\end{equation}
The linguistic term dimension is summed over in the denominator to sample and assign exactly one linguistic term per condition attribute involved in each of the fuzzy logic rules' premises. 
For mathematical ease, transpose the first and second dimensions of ${\varphi (\tilde{\mathbf{I}}')^{T}}$ from Eq.~\ref{eq:restricted_gumbel_dist} so the shape is now $|I_{\mathcal{C}}| \times |U| \times {\max_{i \in I_{\mathcal{C}}}(|\mathcal{M}_{i}|)}$. The rest of this differentiable sampling is performed such that for every $i^{\text{th}}$ condition attribute of fuzzy logic rule $u \in U$, the index, $j$, that maximizes ${\varphi (\tilde{\mathbf{I}}')}_{i, u, j}$ (i.e., \texttt{argmax} of last dimension) is selected; this forms a one-hot-encoding where $\hat{\mathbf{I}}_{i, u, j} = 1$ if the $j^{\text{th}}$ term for the $i^{\text{th}}$ condition attribute was selected for the ${u}^{\text{th}}$ fuzzy logic rule, else $\hat{\mathbf{I}}_{i, u, j} = 0$. Due to these calculations, only one term is selected per condition attribute, per fuzzy logic rule \textemdash{} as desired. Finally, to facilitate the differentiability of obtaining $\mathbf{I}$, a computational trick is employed that uses $\mathbf{I}$ and $\varphi (\tilde{\mathbf{I}}')$ in the forward and backward passes, respectively \cite{bengio2013estimatingpropagatinggradientsstochastic}: 
\begin{equation}
\label{eq:ste}
    \mathbf{I} = \big ( \varphi (\tilde{\mathbf{I}}') + {(\hat{\mathbf{I}}' - {\varphi (\tilde{\mathbf{I}}'))}_\texttt{detached}} \big )^{T}
\end{equation}
where 
$\hat{\mathbf{I}}$ is the one-hot $\texttt{argmax}$: 
\begin{equation}
    \hat{\mathbf{I}}_{i, u, j} = 
    \begin{cases} 
        0 & \varphi(\tilde{\mathbf{I}})_{i, u, j} < \max\limits_{j'}(\varphi(\tilde{\mathbf{I}})_{i, u, j'}) \\
        1 & \text{otherwise} \\
    \end{cases}
\end{equation}

Thus, $\mathbf{I}$ is reconstructed and represents connections between input terms and the rules layer (i.e., which terms form the fuzzy logic rules' conditions). Observe the difference between STGE and STE would only amount to whether Eq.~\ref{eq:ste} uses ${\varphi (\tilde{\mathbf{I}}')^{T}}$ (STGE) or $\tsup[1]{\mathbf{I}}$ (STE \textemdash{} skip Eq.~\ref{eq:add_noise_soften_distribution} and \ref{eq:restricted_gumbel_dist}). 
Either process may facilitate the differentiable generation of fuzzy logic rules. To the best of our knowledge, this is the first NFN to be fully compatible with arbitrary neural architectures, such as CNNs, similar to DNNs. This is possible by its flexibility with random initialization as well as its ability to construct, adapt, and redesign itself purely from gradients. 



\subsection{Batch-Delayed Neurogenesis}\label{section:med}

While it is often desirable for NFNs to be $\epsilon$-complete to match each input to a fuzzy set of at least $\epsilon$ or greater, we observed \textit{thrashing} of the internal connections between the premise and rule layers if new fuzzy sets were frequently added during our preliminary experimentation. Again, this is because GBNA must reassess the conditions involved in the fuzzy rule base (FRB) to explore whether there is an advantage to using this new fuzzy set in the FRB, and if so, for which fuzzy logic rules. 
This naturally led to our hypothesis that this constant thrashing caused by adding a new fuzzy set may result in poor or unstable performance, as the GBNA may no longer continue converging to a particular structure (i.e., FRB), resulting in NFN's instability. This motivated our investigation into how neurogenesis can be safely implemented in an NFN by exploring two approaches: adding the fuzzy set (1) \textit{immediately} or (2) after some \textit{delay}. In essence, while $\epsilon$-completeness is often necessary \textit{in theory}, we wanted to examine how critical it was \textit{in practice}. Furthermore, it is also unclear how robust GBNA is when new fuzzy sets are dynamically added. 

By comparing the two approaches, immediately or delaying adding the new fuzzy set, it is possible to determine whether our preliminary concerns were legitimate or if GBNA can quickly recover. More broadly, this indicates whether STGE (or STE) can handle the expansion of their corresponding categorical distributions inside a neuro-symbolic architecture (in this case, an NFN). 
If adding fuzzy sets immediately leads to no loss in performance, then concerns regarding internal thrashing during training can potentially be dismissed. Otherwise, 
it indicates that this sporadic thrashing may prevent convergence, but delaying neurogenesis alleviates this dilemma.


For these reasons, we propose \textit{batch-delayed neurogenesis} of the membership layer to satisfy $\epsilon$-completeness while alleviating internal thrashing simultaneously. For this paper, only Gaussian membership functions are used to describe fuzzy sets so that helpful computational tricks may be leveraged (as discussed in~\S~\ref{appendix:computational_tricks} and~\ref{section:CoD}); thus, when a new term is to be created, it relies upon selecting a reasonable standard deviation, $\sigma$. Random selection might work in some settings, but may fail across various application domains. If $\sigma$ is too small, the newly added fuzzy set will never be activated. If it is too large, it may catastrophically dominate its peers for that condition attribute by always overcoming all other terms' memberships. Thoughtful initialization of $\sigma$ is the key to maintaining this delicate balance. 

A more reasonable $\sigma$ for a new fuzzy set might be obtained by postponing fuzzy set creation and instead iteratively calculating $\sigma$ on batches of observations; this also has the added benefit of mitigating the issue of FRB thrashing. Batch-delayed neurogenesis executes the following procedure:
\begin{enumerate}
    \item For each observation, find condition attributes that fail $\epsilon$-completeness.
    \item Iteratively calculate problematic condition attributes' means and their variances by \textit{Welford's method for computing variance} \cite{Welford1962NoteOA}. 
\end{enumerate}
Welford's method iteratively calculates the mean and \emph{variance} as new data is received. This is crucial when delaying adding new terms, as it saves memory from having to store all the relevant data; additionally, it helps prevent numerical overflow from naïve calculation. Once enough batches are processed (a hyperparameter), the new fuzzy sets can be created using the means and the square root of the variances for their centers and widths, respectively. Delaying the addition of new fuzzy sets allows the membership layer to learn the stimuli domain and helps filter out atypical values or noise.

\subsection{Encouraging Sparsity in Firing Levels}
Recognizing that the \texttt{softmax} formula (Eq.~\ref{eq:softmax} or Eq.~\ref{eq:norm_softmax}) saturates firing strengths of the NFN's FRB, we began to consider equivalent substitutions to \texttt{softmax} for NFN inference. In particular, one issue with \texttt{softmax} is that it will not assign a zero value to an entry despite insufficient support. In determining FRB applicability, truncating a fuzzy logic rule's activation would be advantageous if the value is extremely close to zero, as it is almost negligible but technically still affects the output. Then, a human might gain more significant insights into the NFN's decision-making by focusing only on a handful of rules instead of the entire FRB. 
One such substitution, \texttt{sparsemax} \cite{10.5555/3045390.3045561}, truncates near-zero values to become zero but may be too aggressive in dismissing FRB's firing levels (found during our personal but unreported exploratory experimentation). On the contrary, $\alpha$\texttt{-entmax} is a family of transforms that represents a generalization of both \texttt{softmax} (when $\alpha$=1) and \texttt{sparsemax} (when $\alpha$=2) \cite{peters-etal-2019-sparse}. 
The $\alpha$\texttt{-entmax} pushes near-zero values to zero for sparse attention in favor of the most critical and strongly activated fuzzy logic rules while simultaneously offering possibilities for better interpretability (e.g., with 256 rules, only three might be activated with no loss in performance). If \texttt{softmax} may be swapped with $\alpha$\texttt{-entmax}, this could greatly benefit NFNs with large FRBs. In this work, only $\alpha=1.5$ is considered for $\alpha$\texttt{-entmax} (i.e., a tradeoff between \texttt{softmax} and \texttt{sparsemax}). 

\subsection{Amplifying Small Differences in Firing Levels}
\label{section:CoD}
Calculating the applicability of a fuzzy logic rule with many conditions may cause numerical underflow in an NFN \cite{cui_curse_2021}. 
More specifically, \CuiEtAl~reveal how \texttt{softmax}'s presence in TSK NFNs saturates fuzzy logic rules' activations \cite{cui_curse_2021}. 
Thus, it is possible for the final prediction to be dominated by a single fuzzy logic rule as the number of condition attributes increases. This is likely undesirable as solely relying on a single fuzzy logic rule for prediction may result in fragile behavior (e.g., overfitting). \CuiEtAl's suggest mitigating this by substituting the summation in Eq.~\ref{eq:mu_rule_sum} with the mean
\begin{equation}
    \label{eq:mu_rule_mean}
    w_{u}(\mathbf{x}) = - \frac{1}{|\mathcal{C}|} \sum\limits_{i \in I_{\mathcal{C}}} 
    \frac{({x}_{i} - {c}_{i, j, u})^2}{2\sigma_{i, j, u}^2} .
\end{equation}
\noindent
\CuiEtAl~argue that the scale of $w_{u}(\mathbf{x})$ (by Eq.~\ref{eq:mu_rule_mean}) 
leads to more stable performance. In this paper, all NFNs will use some form of this computational trick in their implementation; every NFN uses Eq.~\ref{eq:norm_softmax}, but some may use the mean (Eq.~\ref{eq:mu_rule_mean}), whereas others will stay with the standard sum (Eq.~\ref{eq:mu_rule_sum}). 

However, we present unique commentary regarding \CuiEtAl's observations. 
Although \CuiEtAl's Eq.~\ref{eq:mu_rule_mean} elevates the firing level of fuzzy logic rules, it unfortunately neutralizes the discernibility between their activations. 
The \texttt{softmax} not only becomes saturated by rule activations approaching zero due to a rise of input dimensionality but also saturates rule activations as fuzzy logic rule count (i.e., $|U|$) increases when it forces all fuzzy logic rules' activation to sum to 1. 
Consequently, rule activations are pushed to zero from two directions: (1) input dimensionality and (2) fuzzy logic rule count. Existing work (e.g., \texttt{LogTSK}, \texttt{HTSK} \cite{cui_curse_2021}) attempts to address TSK NFN's curse of dimensionality, but from personal experimentation, these techniques alone are often inadequate as they fundamentally fail to scale well as $|U|$ increases. 

We propose altering Eq.~\ref{eq:softmax} to remedy the relative fuzzy logic rules' strengths from becoming suppressed by many poorly activated fuzzy logic rules; specifically, we alleviate issues stemming from the curse of dimensionality concerning fuzzy logic rules found in TSK NFNs by balancing
fuzzy logic rule activations with \textit{layer normalization} \cite{ba2016layer}. The layer normalization is applied \textit{before} the exponential function, such that normalization statistics are computed across all unnormalized fuzzy logic rules:

\begin{minipage}{0.45\linewidth}
\begin{equation}
    \Bar{\mu} = \frac{1}{|U|} \sum_{u \in U} {w}_{u}(\mathbf{x})
\end{equation}
\end{minipage}
\begin{minipage}{0.45\linewidth}
\begin{equation}
    \Bar{\sigma} = \sqrt{\frac{1}{|U|} \sum_{u \in U} \big( {w}_{u}(\mathbf{x}) - \Bar{\mu} \big)^{2}}
\end{equation}
\end{minipage}

\noindent where all fuzzy logic rules share the calculated $\Bar{\mu}$ and $\Bar{\sigma}$. Incorporating layer normalization with the preliminary fuzzy logic rule activation is achieved by
\begin{equation}\label{eq:proposed_norm_rule}
        {w}_{u}'(\mathbf{x}) = {
    \frac{
        w_{u}(\mathbf{x}) - \Bar{\mu}}{\Bar{\sigma}} * \kappa +\beta}
    \end{equation}
where $\kappa$ and $\beta$ are learnable affine transform parameters, and $\Bar{\sigma}$ is calculated via biased estimator. In essence, by re-scaling and re-centering the fuzzy logic rule activations, the relative differences can be enhanced (Eq.~\ref{eq:proposed_norm_rule}) even if the fuzzy logic rules were weakly activated, as they likely are in a scenario with significant dimensionality. Subtracting the max fuzzy logic rule value before applying $\exp$ in Eq.~\ref{eq:norm_softmax} is required to coerce all values within $(-\inf, 0]$. Finally, ${w}_{u}'$ may be used in place of ${w}_{u}$ for Eq.~\ref{eq:norm_softmax}. The various interactions between these computational tricks, firing level enhancements, and amplifications are explored in the experiments. 
~\ref{appendix:firing_levels} contains visual aids.

\section{A First-Person Shooter Video Game}\label{reward_scenarios:doom}
Although Atari 2600 games are often used as benchmarks for vision-based tasks involving RL \cite{mnih2015humanlevelcontrol}, they represent non-realistic two-dimensional environments from a third-person perspective \cite{kempka2016vizdoom}. Instead, this paper will investigate how the proposed concurrent optimization of NFNs behaves when given visual stimuli from a first-person perspective in a ``semi-realistic'' three-dimensional world. These three-dimensional scenarios are offered through a library called ViZDoom \cite{kempka2016vizdoom, Wydmuch2019ViZdoom}, which is based on the classic first-person shooter (FPS) game called DOOM. 

Evaluating artificial intelligence (AI) research on DOOM is particularly compelling as it has several advantages. DOOM is open source via 
GNU General Public License, written in C++, compatible with Windows/Linux/Mac operating systems, offers small resolution (i.e., < 640x480), has low system requirements, requires little disk space (approximately forty megabytes), has an active community and steady brand recognition \cite{kempka2016vizdoom}. The performance of ViZDoom is remarkable. It only uses a single core from a central processing unit and can render 7,000 low-resolution frames per second on a Linux machine running an Intel Core i7-4790k \cite{kempka2016vizdoom}.

ViZDoom was chosen as the primary evaluation environment to demonstrate the concurrent optimization of NFNs for the following reasons:
\begin{enumerate}
    \item To showcase concurrent optimization is the first technique to construct NFNs in online vision-based RL tasks. To our knowledge, no existing NFN research can dynamically build an NFN in response to how well it performs in a vision-based scenario with RL.
    \item ViZDoom offers challenging scenarios, which are great for highlighting concurrent optimization's ability to handle complex challenges.
    \item ViZDoom is very efficient concerning computation and memory, allowing for straightforward, large-scale experimentation.
\end{enumerate}

\subsection{Basic Weapon Usage}
There are three scenarios about basic weapon usage \textemdash{} they are concerned with handling pistols or rocket launchers. The first two scenarios are relatively simple, but the third scenario is more challenging as it requires a greater understanding of ballistics and enemy movement. Each scenario of this section will use enemy monsters as targets to practice the handling of weapons. Still, the enemy monsters will not attempt to fight the player yet (that will occur in the other scenarios where they are present).

\begin{figure*}[ht]
    \centering
    \begin{minipage}{0.4\linewidth}
    \centering
    \textit{(1) Aiming Pistol}\\
    \includegraphics[width=.85\linewidth, trim={60bp 38bp 50bp 38bp}, clip]{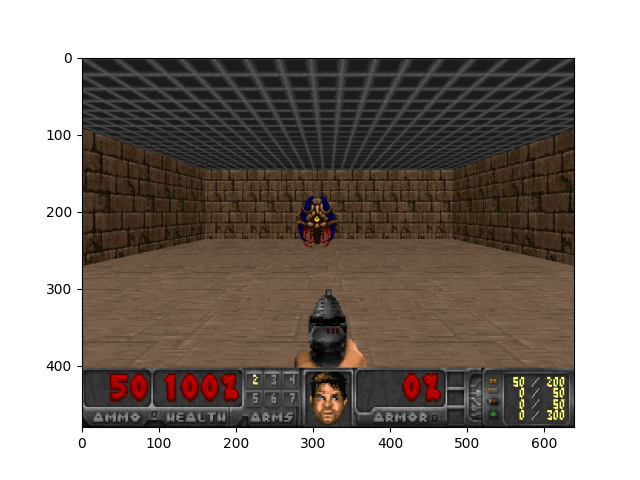}
    \end{minipage}
    \begin{minipage}{0.4\linewidth}
    \centering
    \textit{(2) Firing Rocket Launcher}\\
    \includegraphics[width=.85\linewidth, trim={60bp 38bp 50bp 38bp}, clip]{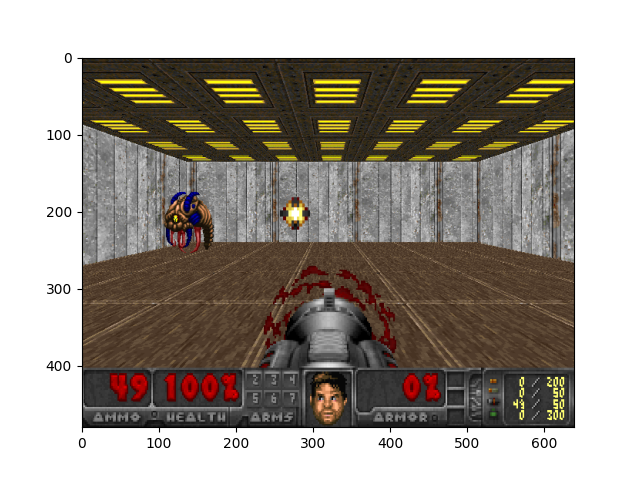} 
    \end{minipage}
    \caption{Attacking stationary targets with a pistol (left) and a rocket launcher (right).}
    \label{fig:doom_basic}
\end{figure*}

\subsubsection{Target Practice with Pistol}
The first scenario involving target practice with a pistol is relatively simple. The primary purpose of this scenario is to check whether training the given AI agent in a three-dimensional environment is feasible. 
The player is spawned inside a rectangular room with brown walls and floor but a gray ceiling. They are positioned along the longer wall in the center. A monster is randomly spawned somewhere along the opposite wall but cannot move or fight (it may only look around). The player can only go left, right, or shoot. A single hit is enough to kill the monster, but the player is given fifty rounds of ammunition. The episode finishes when the monster is killed or on timeout. Refer to Figure \ref{fig:doom_basic}.1 for an example initial state showing the player aiming the pistol at the enemy monster. The worst possible score is $-300$.

\subsubsection{Handling Rocket Launchers}
Target practice with the rocket launcher follows the same environmental rules as the pistol target practice scenario, but the room's textures differ. Specifically, this room has a brown floor, light gray walls, and yellow square tiles arranged across a dark gray ceiling. Learning to use the rocket launcher is slightly more challenging than using the pistol, as the projectile's eventual impact is delayed. For example, in Figure \ref{fig:doom_basic}.2, the player fires a rocket launcher toward the enemy monster, but the projectile inevitably misses. The player is provided fifty rockets; the worst possible score is $-300$.

\subsubsection{Target Tracking \& Trajectory Prediction}
\begin{figure*}[ht]
    \centering
    \begin{minipage}{0.4\linewidth}
    \centering
    \textit{(1) Initial State}\\
    \includegraphics[width=.85\linewidth, trim={60bp 38bp 50bp 38bp}, clip]{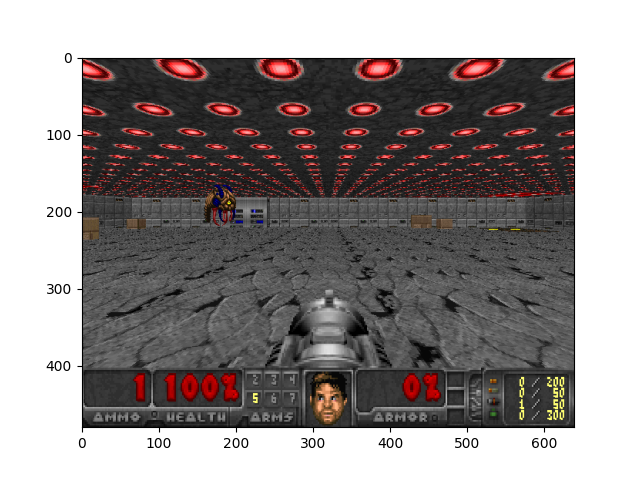}
    \end{minipage}
    \begin{minipage}{0.4\linewidth}
    \centering
    \textit{(2) Player Attacking}\\
    \includegraphics[width=.85\linewidth, trim={60bp 38bp 50bp 38bp}, clip]{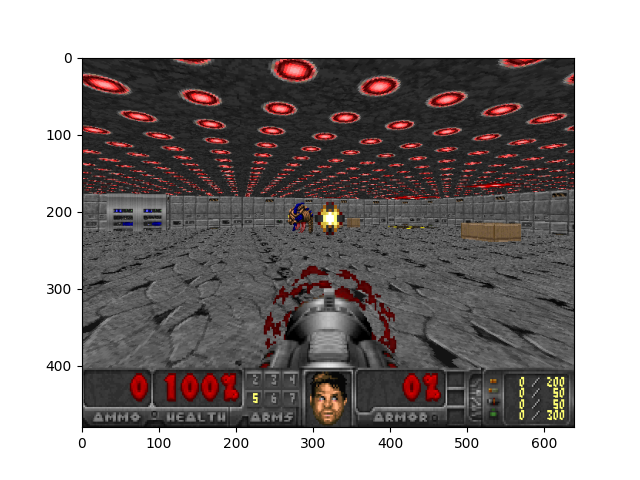} 
    \end{minipage}
    \caption{Tracking an enemy and predicting their trajectory to launch an attack.}
    \label{fig:doom_predict_position}
\end{figure*}

The scenario for the basic usage and handling of a rocket launcher considers only a stationary enemy monster. Realistically, enemy monsters will be moving around \textemdash{} either toward the player or attempting to evade the player's attacks. In this scenario, the player is provided a rocket launcher with a single projectile, and an enemy monster (same one as the other scenarios) is randomly spawned somewhere along the opposite wall. This enemy monster moves along the wall between the left and right corners. The player may turn left, turn right, or fire their rocket. If they kill the monster, they receive a reward of +1 but are punished for every time step the player is alive by -0.0001. The episode ends after 300 time steps, the monster is killed, or the missile hits a wall. This is the most challenging scenario of the three weapon handling tasks, requiring the player to \textit{predict position}. Thus, an AI agent must then learn how to track an enemy monster, estimate its current trajectory, and determine an interception point for the rocket launcher's projectile to hit its target. Figure~\ref{fig:doom_predict_position} illustrates how the enemy monster may move around as the player attempts to launch a rocket toward it. 

\subsection{Stand One's Ground}
Two scenarios require the player to hold their ground for as long as possible as they cannot move \textemdash{} they can only look left, look right, or fire their weapon (a pistol). Health for the player begins at 100\%, and there is no possibility to obtain more health. The player is rewarded for killing enemy monsters. In the most difficult of the two scenarios, the player is provided limited ammunition and must use it sparingly yet appropriately when necessary. Each episode ends upon the player's death, which is inevitable due to the overwhelming enemy difficulty, or possibly due to ammunition depletion.

\subsubsection{Hold The Line}
\begin{figure*}[ht]
    \centering
    \begin{minipage}{0.4\linewidth}
    \centering
    \textit{(1) Initial State}\\
    \includegraphics[width=0.85\linewidth, trim={60bp 38bp 50bp 38bp}, clip]{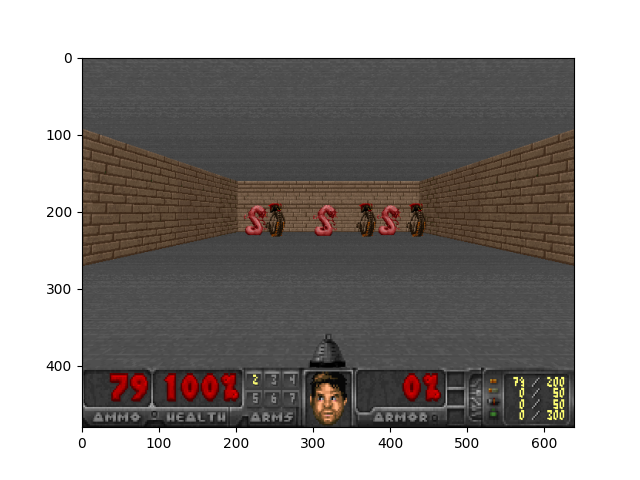}
    \end{minipage}
    \begin{minipage}{0.4\linewidth}
    \centering
    \textit{(2) Player Attacking}\\
    \includegraphics[width=0.85\linewidth, trim={60bp 38bp 50bp 38bp}, clip]{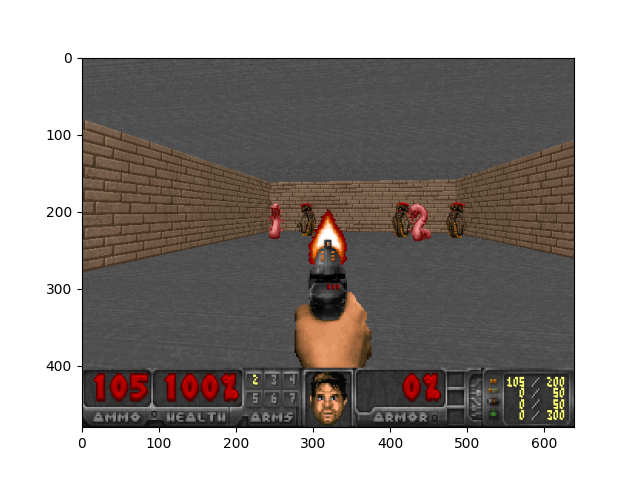}
    \end{minipage}
    
    \begin{minipage}{0.4\linewidth}
    \centering
    \textit{(3) Enemy Attacking}\\
    \includegraphics[width=0.85\linewidth, trim={60bp 38bp 50bp 38bp}, clip]{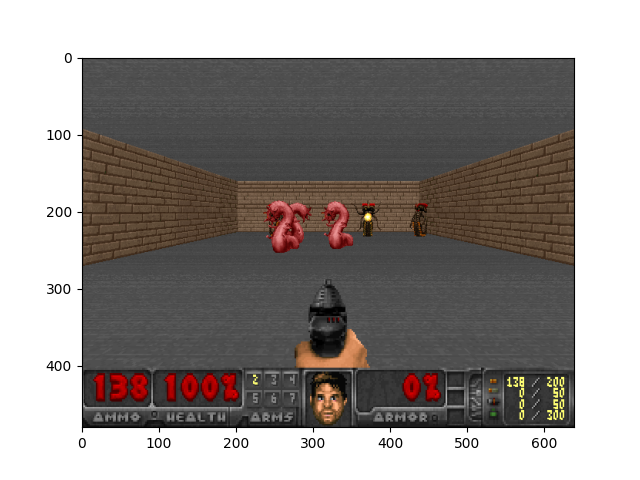}
    \end{minipage}
    \begin{minipage}{0.4\linewidth}
    \centering
    \textit{(4) Player Taking Damage}\\
    \includegraphics[width=0.85\linewidth, trim={60bp 38bp 50bp 38bp}, clip]{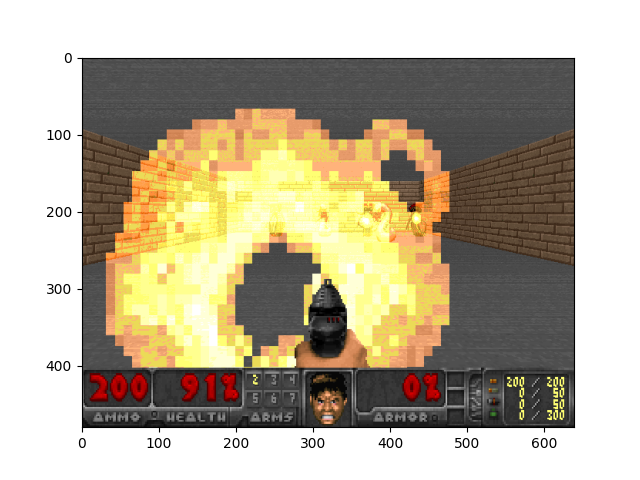}
    \end{minipage}
    \caption{Holding the line against a horde of enemy monsters.}
    \label{fig:doom_defend_the_line}
\end{figure*}

The first scenario, shown in Figure~\ref{fig:doom_defend_the_line}, forces the player to \textit{hold the line} from increasingly more difficult enemies than the last. The player spawns centered along the shorter wall of a rectangular room; three melee-only and three shooting monsters will then spawn along the opposite wall (Figure~\ref{fig:doom_defend_the_line}.1). At first, a single shot can defeat each enemy (Figure~\ref{fig:doom_defend_the_line}.2). However, they gradually become more resilient upon respawning after a brief delay. The player is given infinite ammunition, but the ammo counter in the HUD increases until two hundred rounds are shown. The fireball projectiles launched by the enemy monsters (Figure~\ref{fig:doom_defend_the_line}.3) will always hit the player (Figure~\ref{fig:doom_defend_the_line}.4) unless a melee-only monster interrupts its path; should that occur, the fireball projectile will kill the melee-only enemy. This scenario is challenging for an AI agent to learn as it requires shifting focus between enemy types. In particular, a helpful strategy may prioritize killing the enemies capable of launching fireballs until the melee-only enemies become a threat, as it is advantageous to let the melee-only enemies potentially absorb any incoming fireballs.

    

\subsubsection{Defend The Center}
\begin{figure*}[ht]
    \centering
    \begin{minipage}{0.4\linewidth}
    \centering
    \textit{(1) Initial State}\\
    \includegraphics[width=0.85\linewidth, trim={60bp 38bp 50bp 38bp}, clip]{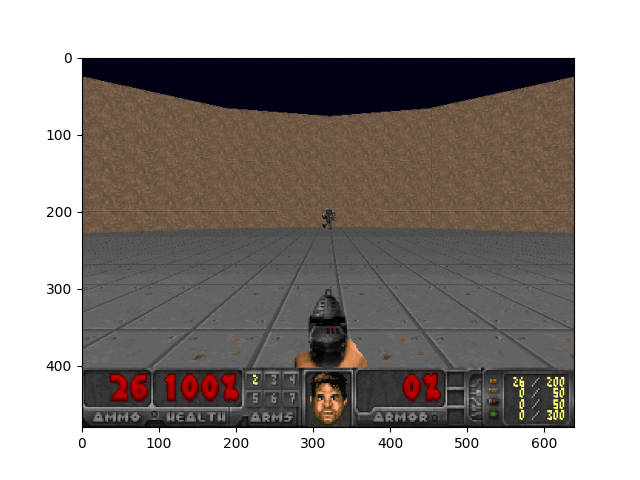}
    \end{minipage}
    \begin{minipage}{0.4\linewidth}
    \centering
    \textit{(2) Player Attacking}\\
    \includegraphics[width=0.85\linewidth, trim={60bp 38bp 50bp 38bp}, clip]{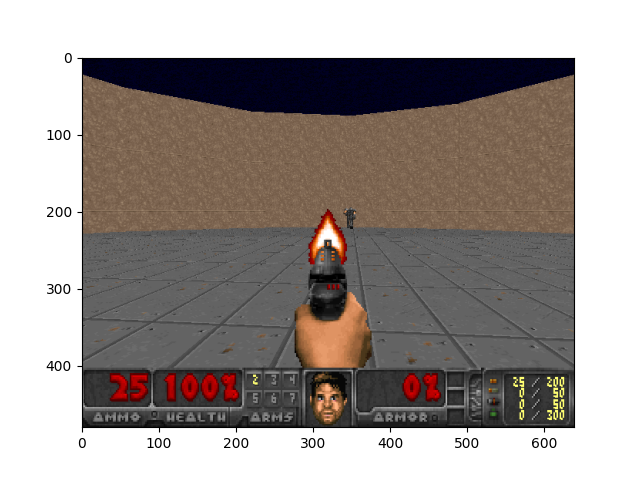}
    \end{minipage}
    
    \begin{minipage}{0.4\linewidth}
    \centering
    \textit{(3) Enemies Flanking}\\
    \includegraphics[width=0.85\linewidth, trim={60bp 38bp 50bp 38bp}, clip]{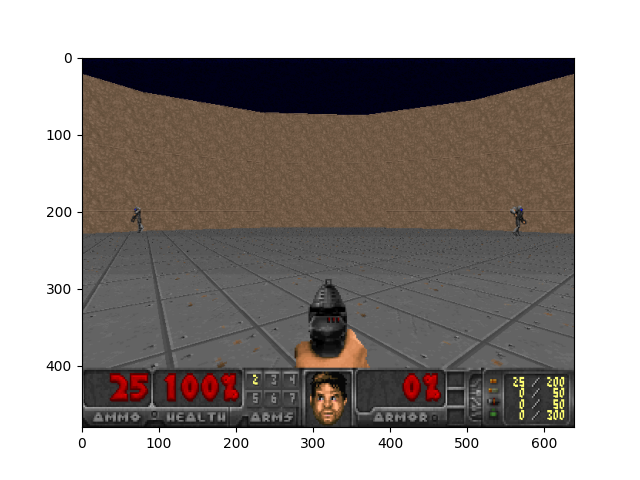}
    \end{minipage}
    \begin{minipage}{0.4\linewidth}
    \centering
    \textit{(4) Facing Imminent Threats}\\
    \includegraphics[width=0.85\linewidth, trim={60bp 38bp 50bp 38bp}, clip]{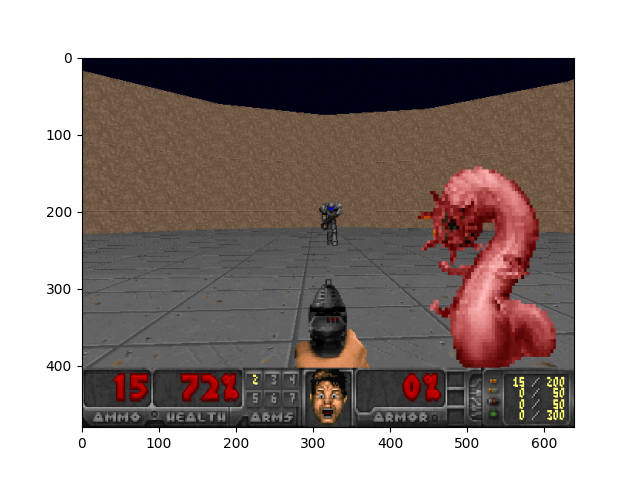} 
    \end{minipage}

    \caption{Handling imminent threats while balancing situational awareness.}
    \label{fig:doom_defend_the_center}
\end{figure*}

The second scenario places the player at the center of a large circular room, where enemies spawn all around them, and the player needs to \textit{defend the center}. Both melee-only and shooting enemies will try to attack the player, but the shooting enemies can only fire their weapons upon getting closer to the player. The player is given twenty-six rounds of ammunition and cannot acquire more ammunition or health. However, the monsters are killed after a single shot but will later spawn again after some time. The player receives a $+1$ reward for every monster they defeat and a $-1$ for their eventual death. Therefore, this scenario's worst and best possible scores are $-1$ and $25$, respectively. Figure~\ref{fig:doom_defend_the_center} depicts a player in this scenario. 


While both scenarios are challenging, defending the center is possibly more difficult than holding the line, as the player must learn to defeat enemy threats while preserving ammunition and maintaining situational awareness by watching their surroundings. In particular, the player can be flanked by enemies and receive damage from monsters off-screen. Thus, an agent must learn to recognize off-screen threats while also determining if on-screen enemies are an immediate threat. 

\subsection{Route-Finding \& Evasion}
All scenarios discussed thus far have prohibited players from moving through or navigating their environment. If the player could move through their environment, this may allow them to develop more successful strategies. For instance, defending the center would become less challenging as the player could move out of the enemy monster's attack range. However, by allowing the player to navigate their environment, even more difficult scenarios may be given for them to master, such as gathering health over a pool of acid or dodging relentless enemy attacks.

\subsubsection{Gathering Health}
\begin{figure*}[ht]
    \centering
    \begin{minipage}{0.4\linewidth}
    \centering
    \textit{(1) Initial State}\\
    \includegraphics[width=0.85\linewidth, trim={60bp 38bp 50bp 38bp}, clip]{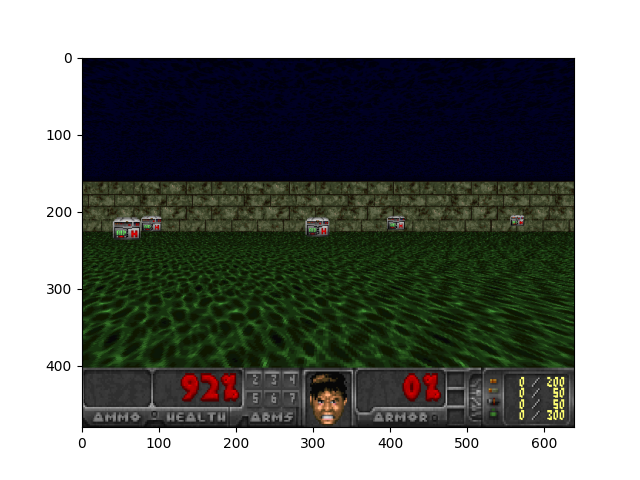}
    \end{minipage}
    \begin{minipage}{0.4\linewidth}
    \centering
    \textit{(2) Medkits Dropped}\\
    \includegraphics[width=0.85\linewidth, trim={60bp 38bp 50bp 38bp}, clip]{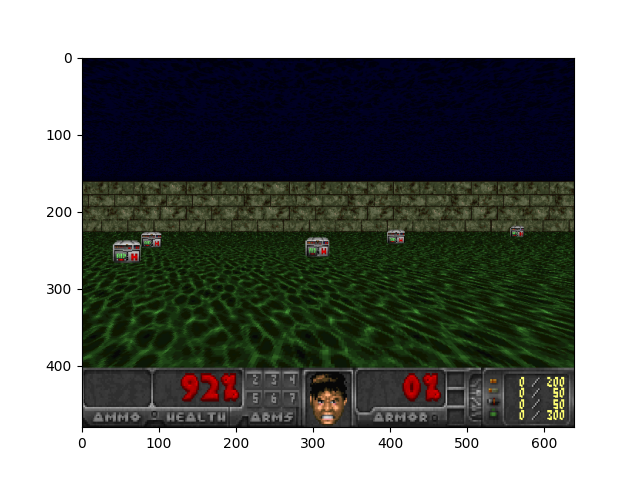}
    \end{minipage}
    
    \begin{minipage}{0.4\linewidth}
    \centering
    \textit{(3) Player Moving}\\
    \includegraphics[width=0.85\linewidth, trim={60bp 38bp 50bp 38bp}, clip]{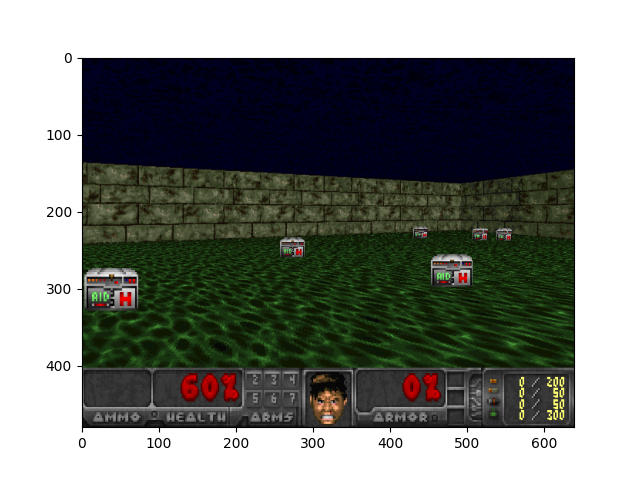}
    \end{minipage}
    \begin{minipage}{0.4\linewidth}
    \centering
    \textit{(4) Player About to Heal}\\
    \includegraphics[width=0.85\linewidth, trim={60bp 38bp 50bp 38bp}, clip]{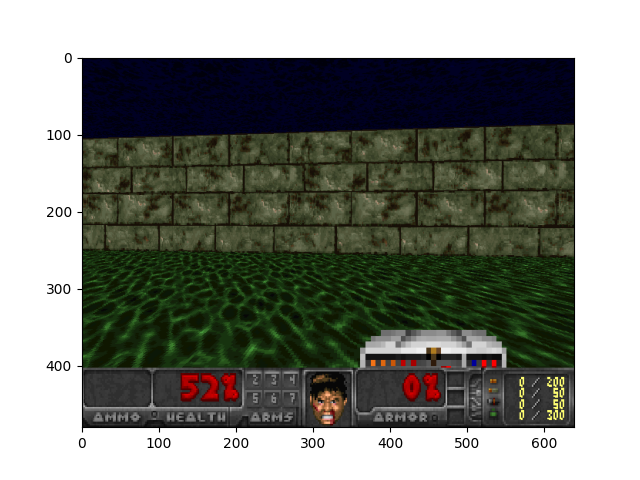} 
    \end{minipage}
    \caption{Surviving a room with an acidic floor by gathering health.}
    \label{fig:doom_health_gathering}
\end{figure*}

The player is trapped in a rectangular room filled with a green, acidic floor, which damages the player over time (Figure~\ref{fig:doom_health_gathering}.1). Medkits fall from the sky and are initially spread uniformly over the map (Figure~\ref{fig:doom_health_gathering}.2). There are three possible actions: move left/right and go forward. The player must \textit{gather health} to survive this room with an acidic floor (Figure~\ref{fig:doom_health_gathering}.3). A medkit is consumed once the player walks over it (Figure~\ref{fig:doom_health_gathering}.4), and their health increases by $+25\%$ after they collect it. More medkits will continue to appear as they fall from the sky. An AI agent in this scenario must learn to prolong their existence by recognizing that medkits are associated with survival. 
Every time step the player survives, they receive a +1 reward, but if they die, then a punishment of -100 is given. 


\subsubsection{Take Cover}
\begin{figure*}[ht]
\centering
\begin{minipage}{0.4\linewidth}
\centering
\textit{(1) Initial State}\\
\includegraphics[width=0.85\linewidth, trim={60bp 38bp 50bp 38bp}, clip]{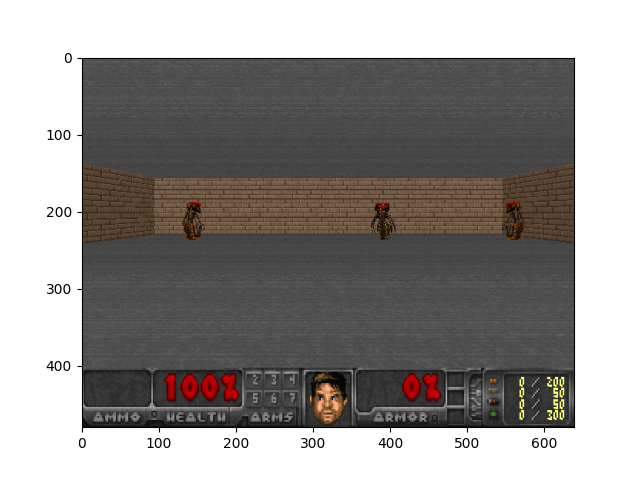}
\end{minipage}
\begin{minipage}{0.4\linewidth}
\centering
\textit{(2) Enemies Begin Attack}\\
\includegraphics[width=0.85\linewidth, trim={60bp 38bp 50bp 38bp}, clip]{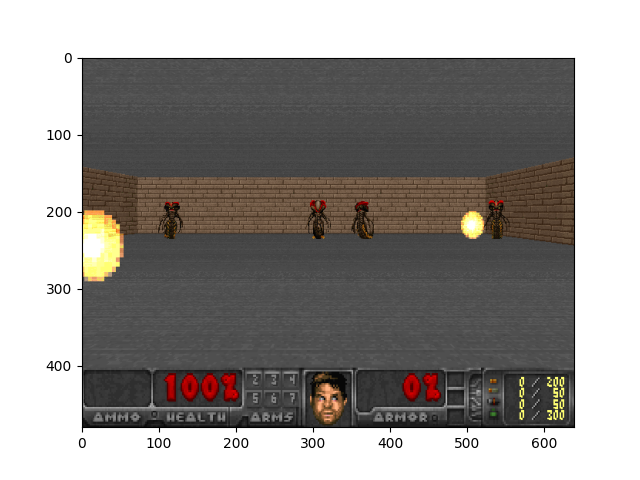}
\end{minipage}

\begin{minipage}{0.4\linewidth}
\centering
\textit{(3) Danger Increases}\\
\includegraphics[width=0.85\linewidth, trim={60bp 38bp 50bp 38bp}, clip]{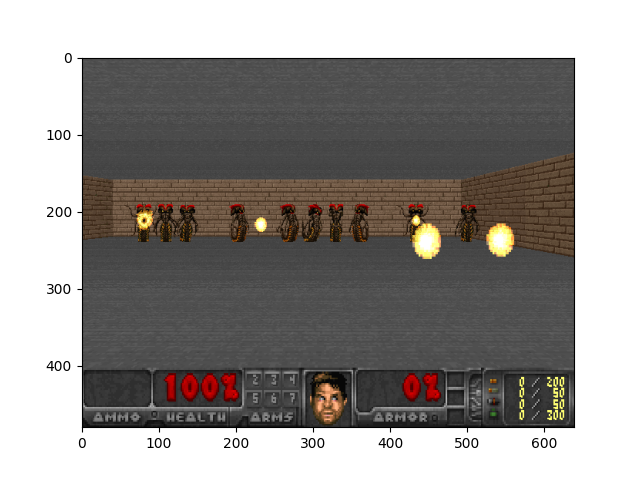}
\end{minipage}
\begin{minipage}{0.4\linewidth}
\centering
\textit{(4) Player Taking Damage}\\
\includegraphics[width=0.85\linewidth, trim={60bp 38bp 50bp 38bp}, clip]{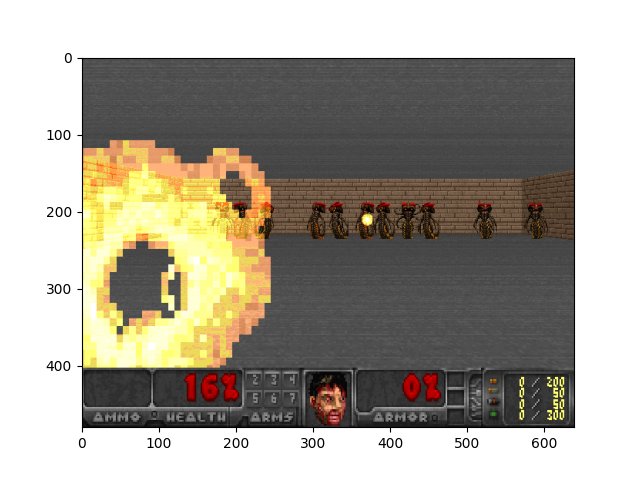}
\end{minipage}

\caption{Dodging a relentless onslaught of fireballs from enemy monsters.}
\label{fig:doom_take_cover}
\end{figure*}

This scenario is similar to where the player must hold the line, but they are centered along the longest wall this time. Across from the player, enemy monsters randomly spawn that can shoot fireballs. The player must dodge their attacks by moving left or right, but more enemies will appear over time. Death is inevitable, and an episode will end once the player dies. A reward of $+1$ is given for every time step the player survives as they \textit{take cover}. An example sequence of how this scenario plays out is shown in Figure~\ref{fig:doom_take_cover}. 

\section{Playing DOOM}
We benchmark concurrent optimization of NFNs in a vision-based FPS video game called DOOM by training the structure and parameters of hierarchical NFNs with an online RL algorithm. Agents must learn how to use weaponry, withstand enemy hordes, take cover, and gather health. Section~\ref{reward_scenarios:doom} provides sample images of these scenarios and describes them in detail. Concurrent optimization of NFNs requires no modifications to the online RL algorithm and remains applicable to other learning paradigms. Moreover, concurrent optimization of NFNs allows them to be randomly initialized and utilized similarly to DNNs. For experiments, we primarily focused on online RL since existing literature suggests prior work is generally incapable of simultaneously training NFN's structure and parameters (Section~\ref{section:related_work}). Typically, prior work conducted in training NFNs with online RL would only focus on the parameters, where the structure (i.e., fuzzy logic rules) was given, or assumed, based on a priori expert knowledge. 

\subsection{Experiment Setup}
Since DOOM is an FPS video game, each (unaltered) state of the environment is an RGB24 image with 640x480 resolution. The available actions, $\prod_{i \in I_{A}} A_{i}$, are \textit{all possible} combinations of individual actions (e.g., ``do nothing'', ``move left AND fire weapon'', ``move left AND fire weapon AND move right''). Some of these actions, such as ``move left AND fire weapon AND move right'' are possibly redundant as the two moves contradict each other; the agent must learn to ignore such irrelevant actions. Each game scenario followed standard configurations by rendering the heads-up display, weapon, and screen flashes, but not the crosshair, decals, particles, effect sprites, messages, or corpses. Every environment state is preprocessed by normalizing the image data (i.e., dividing by $255$) and resizing the image to 84x84 using pixel area relation for its resampling (i.e., OpenCV's INTER\_AREA). For convenience, a state $\mathbf{s}$ will refer to an image \textit{after} it is preprocessed, and inputs to the agents will expect this type of input. Experiments were performed on an NVIDIA GTX 3080 Graphics Processing Unit (GPU) and each experimental condition comprised of 720 trials.

\subsubsection{Experimental Conditions} 
Due to the inherit difficulty in learning to play DOOM with standard methods such as Deep Q-Learning (DQL) \cite{mnih2015humanlevelcontrol}, agents were trained online with Dueling Double DQL (Duel DDQL) \cite{wang2016dueling, 10.5555/3016100.3016191}, such that their underlying functions (e.g., state value, advantage) are either represented by a DNN or NFN. To directly compare the neural architectures' performance, two experimental conditions are investigated where \textit{all} functions are either implemented by (1) DNNs or (2) NFNs. Note that it is, of course, possible to mix the two. Hierarchical NFNs will be used in these experimental conditions to mimic the behavior of DNNs. 

\subsubsection{Common Mechanisms \& Processes}
Each experimental condition began with the same rate of exploration ($= 1$) and would decay this rate of exploration by multiplying it with $0.9999$ at the end of each training step; the minimum rate of exploration during training was $0.1$. Input fed into each experimental condition is initially processed by four sequential CNNs (in-channels, out-channels, kernel size, stride) with no bias term in convolution: (1) (3, 8, 3, 2), (2) (8, 8, 3, 2), (3) (8, 8, 5, 1), and (4) (8, 16, 7, 1). The output of every convolution was followed by batch normalization, ReLU, and dropout ($p=0.2$) \cite{JMLR:v15:srivastava14a}. For every input, $\mathbf{s}$, its shape after this sequence of operations was (16, 10, 10). This representation was flattened according to row-major order; let $\mathbf{x}$ refer to this flattened vector such that $|\mathbf{x}| = 1600$. Vectors $\mathbf{x}$ are then fed into a DNN or hierarchical NFN, depending on which experimental condition the agent belongs to. Parameters for the DNN or NFN will be optimized with Adam \cite{adam_optimizer}.

\subsubsection{Mentionable Differences}
DNN's structure will remain \textit{static}, while the NFN's structure is \textit{dynamic} due to concurrent optimization. Thus, only the parameters (e.g., weights, biases) of the DNN will be fine-tuned, whereas the structure and parameters of the NFN are simultaneously fine-tuned. This is possible since the underlying parameters determine NFN's structure (i.e., connections between the condition and rule layers). Future work could explore comparing NFNs with concurrent optimization against DNNs using neural architecture search \cite{zoph2017neuralarchitecturesearchreinforcement} techniques. Still, this paper's primary focus is whether NFNs can even be applied to vision-based tasks using concurrent optimization \textemdash{} not necessarily trying to surpass DNNs, but rather, be comparable to their performance.

\subsection{Policy Induction}
The success of an agent learning to play DOOM may significantly rely upon selecting reasonable hyperparameters. A hyperparameter search was performed with a Bayesian optimization technique called the \textit{Tree Parzen estimator} \cite{10.5555/2986459.2986743}. The Tree Parzen estimator is a model-based approach that uses sophisticated heuristics to search through the hyperparameter space efficiently. Each trial lasted 10 epochs, where each epoch consisted of 500 time steps. The agent was evaluated for 25 episodes at the end of every epoch. The evaluation performance at a given epoch was the average of all episodes' total rewards. Any stochastic behavior of the agent's underlying functions was temporarily suspended during evaluation (e.g., no dropout, or if STGE is used, no sampled noise), and a greedy policy was followed.

\subsubsection{Objective of the Hyperparameter Search}
The evaluation performance of each trial was monitored, with multiple metrics tracking its progression across epochs to quantify its success. More specifically, the multi-objective hyperparameter search must maximize its average (Mean), minimize its standard deviation (SD), and also maximize the slope ($m$) describing the overall trend across each epoch's evaluation performance. In other words, ideal hyperparameters will not only yield a high average performance but also ensure it has little variability and that the agent is continuously or steadily improving (rather than only performing well early on but then on a downward trend); therefore, a hyperparameter configuration with a positive slope suggests that if it were further trained and the trend continued, then greater performance could potentially be witnessed. Selection of the best hyperparameters favored those with the greatest Mean score. If a tie occurred or evaluation performance was too similar, only trials with $m > 0$ were considered, and trials with SD minimized were preferred.

Hyperparameters may correspond to the learning algorithm (e.g., discount factor), whereas others influence environment design or implementation (e.g., skipped frames). 
Given the plethora of design choices in how concurrent optimization of NFNs could be implemented, such as whether to use STE or STGE for the dynamic reconfiguration of the fuzzy logic rules, the hyperparameter search space for NFNs includes these design choices in addition to numerical values.

\subsection{Available Hyperparameters for the Deep Neural Network}\label{appendix:dnn_hyperparameters}
Possible hyperparameters that may be optimized for the DNN and trained by Duel DDQL control the environment, learning algorithm, or architecture. For instance, the environment could only be fine-tuned according to the 
\begin{enumerate}
    \item Skipped frames (both in training and evaluation), ``Frames'': 4, 8, 12
\end{enumerate}
but the Duel DDQL algorithm could be further adjusted as well
\begin{enumerate}
    \item Learning rate, $\eta$: $[1e{-5}, 1e{-3}]$
    \item Size of batch, $|\mathbf{X}|$: $8, 16, 24, 32, 40, 48, 56, 64$
    \item Size of memory for experience replay, ``Mem.'': $10k, 20k, 30k, 40k, 50k$
    \item Discount factor, $\gamma$: $[0.9, 0.99]$
\end{enumerate}

\noindent
The hyperparameter search also determined the DNNs' architecture:

\begin{enumerate}
    \item Number of hidden neurons, $|\mathcal{N}|$: 128, 256, 384, 512
    \item Non-linear activation function, $\mathcal{h}_{\mathcal{n}}$, where $\mathcal{h}_{\mathcal{n}}$ may defined as:
\end{enumerate}
\begin{center}
\scriptsize
    \begin{tabular}{cccccccc}
     \texttt{Sigmoid} & \texttt{Tanh} & \texttt{ReLU} & \texttt{Softshrink} & \texttt{GELU} & \texttt{SELU} & \texttt{Softsign} & \texttt{LeakyReLU} \\
     \texttt{LogSigmoid} & \texttt{Tanhshrink} & \texttt{ReLU6} & \texttt{Hardshrink} & \texttt{SiLU} & \texttt{CELU} & \texttt{Softplus} & \texttt{PReLU} \\
     \texttt{Hardsigmoid} & \texttt{Hardtanh} & & & \texttt{Mish} & \texttt{ELU} & \texttt{Hardswish} & \texttt{RReLU} \\
\end{tabular}
\end{center}

\noindent
for all neurons, $\mathcal{n} \in \mathcal{N}$. Each DNN consisted of two linear layers, each followed immediately by the selected non-linear activation function, $\mathcal{h}_{\mathcal{n}}$.

\subsection{Available Hyperparameters for the Neuro-Fuzzy Network}\label{appendix:nfn_hyperparameters}
Due to the significant number of proposed changes in this paper and to understand their impact, this hyperparameter search would either disable/enable a particular proposed change (e.g., STE or STGE, Eq.~\ref{eq:mu_rule_sum} or Eq.~\ref{eq:mu_rule_mean}) or fine-tune a real-valued hyperparameter (e.g., Gumbel temperature, $\tau$). The hyperparameter search operated over the following options regarding design choices of GBNA, 
\begin{enumerate}
    \item Size of FRB (i.e., count of fuzzy logic rules), $|U|$: 64, 128, 192, 256
    \item Sample fuzzy logic rules with STGE: \texttt{False} (STE) or \texttt{True} (STGE)
    \item Gumbel temperature (STGE only), $\tau$: [0.25, 1.25]
    \item Threshold (STGE only), $\theta$: 0 (disabled) or 10 (slightly enabled)
    \item Retain sampled Gumbel noise, $\mathbf{N}$, after \makebox[1cm]{\hrulefill} batches: 1 (i.e., sample new noise each forward pass), 32, 64, 128, 256
\end{enumerate}

\noindent
delayed neurogenesis,
\begin{enumerate}
    \item Membership degree that must be reached for each input's attributes across their respective terms (i.e., $\epsilon$-completeness), $\epsilon$: [0.1, 0.5]
    \item Batches to wait before adding the new fuzzy set being built by Welford's method, $+\mu$: 1 (i.e., immediately add the new fuzzy set), 3, 5 
\end{enumerate}

\noindent
or how to address the curse of dimensionality:
\begin{enumerate}
    \item Preliminary calculation of fuzzy logic rule's applicability, ${w}_{u}$: \texttt{Sum} (Eq.~\ref{eq:mu_rule_sum}) or \texttt{Mean} (\CuiEtAl's Eq.~\ref{eq:mu_rule_mean})
    \item Activation of fuzzy logic rule, $\overline{w}_{u}$, when Eq.~\ref{eq:norm_softmax} is generalized to $\alpha$-\texttt{entmax}, where $\alpha$: 1.0 (i.e., \texttt{softmax}) or 1.5 (i.e., 1.5-\texttt{entmax})
    \item Incorporate certainty factors (CFs) \cite{kim_hyfis_1999}, CF: \texttt{False} or \texttt{True}
    \item Enable layer normalization (LN), LN: \texttt{False} or \texttt{True}
\end{enumerate}

\noindent
This hyperparameter search also fine-tuned: 
\begin{enumerate}
	\item Learning rate, $\eta$: $[1e{-4}, 1e{-3}]$
\end{enumerate}

Other hyperparameters regarding the environment or learning algorithm were determined by the DNN's corresponding best trial. Investigating each possible combination with grid search hyperparameter tuning would have been computationally intractable. Instead, the Tree Parzen estimator efficiently searches through possibilities by relying on historical trials to place greater attention on tuning hyperparameters shown to be more important. 

\subsection{Experimental Results}
Hyperparameter abbreviations (in \S~\ref{appendix:dnn_hyperparameters} and \ref{appendix:nfn_hyperparameters}) are used for conciseness. Results of the best hyperparameters discovered by the Tree Parzen estimator search for DNN and NFN are in Tables~\ref{tab:doom_dnn_best_hyperparameters} and \ref{tab:doom_nfn_best_hyperparameters}, respectively. Table~\ref{tab:doom_nfn_best_hyperparameters} omits the hyperparameter for \textit{constraining} STGE, $\theta$, as it was ineffective across all scenarios (refer to~\ref{appendix:constraint}). 
For each task, performance discrepancies between DNN and NFN will be analyzed by the appropriate two-sample t-tests based on whether equality of variances can reasonably be assumed.

\begin{table}[ht]
    \centering
    \addtolength{\tabcolsep}{-0.3em}  
    \begin{tabular}{c||c|c|c|c|c|c|c||c|c|c}
        Task & $|\mathcal{N}|$ & $\mathcal{h}_{\mathcal{n}}$ & $\eta$ & $|\mathbf{X}|$ & Mem. & $\gamma$ & Frames & Mean & SD & $m$ \\
        \hline\hline

        BP & 512 & \texttt{Hardtanh} & 6.9e-4 & 56 & 30k & .95 & 8 & 85.40 & 9.33 & \textbf{26.5}  \\ 

        BR & 256 & \texttt{Hardswish} & 9.7e-4 & 48 & 10k & .97 & 12 & \textbf{58.40} & \textbf{10.0} & \textbf{19.6}  \\ 

        PP & 512 & \texttt{LogSigmoid} & 8.2e-5 & 56 & 30k & .94 & 4 & 0.29 & \textbf{0.48} & \textbf{0.04}  \\ 

        \hline
        
        HTL & 256 & \texttt{RReLU} & 5.0e-4 & 48 & 20k & .97 & 4 & \textbf{27.24} & 5.57 & \textbf{1.98} \\
        DTC & 128 & \texttt{ReLU} & 3.3e-4 & 16 & 40k & .90 & 8 & \textbf{17.92} & \textbf{1.96} & \textbf{1.12} \\

        \hline

        TC & 128 & \texttt{LogSigmoid} & 5.9e-4 & 24 & 30k & .94 & 12 & 481.1 & 297 & 4.22  \\ 

        GH & 128 & \texttt{RReLU} & 3.7e-5 & 16 & 30k & .94 & 12 & \textbf{2100} & \textbf{0.0} & 29.9  \\ 
        
    \end{tabular}
    \caption{Best trials' hyperparameters for DNNs.}
    \label{tab:doom_dnn_best_hyperparameters}
\end{table}
\vspace{-16pt}
\begin{table}[ht]
    \centering
    \addtolength{\tabcolsep}{-0.3em}  
    \begin{tabular}{c||c|c|c|c|c|c|c|c|c|c|c|c||c|c}
        Task & $|U|$ & $w_{u}$ & $\eta$ & $\alpha$ & STGE & CF & LN & $\tau$ & $\epsilon$ &$\mathbf{N}$ & $+\mu$ & Mean & SD & $m$ \\
        \hline\hline

        BP & 192 & \texttt{Sum} & 8.5 & 1.5 & F & T & T & - & .36 & - & 3 & \textbf{86.6} & \textbf{9.24} & 20.3 \\
        
        BR & 256 & \texttt{Sum} & 9.2 & 1.0 & T & T & F & .42 & .35 & 256 & 1 & 37.2 & 41.6 & 16.4 \\  

        PP & 256 & \texttt{Mean} & 8.1 & 1.0 & T & F & F & .52 & .40 & 128 & 1 & \textbf{0.36} & 0.50 & 0.01 \\ 

        \hline

        HTL & 64 & \texttt{Mean} & 1.0 & 1.5 & T & T & F & 1.2 & .40 & 32 & 3 & 25.4 & \textbf{5.43} & 1.59 \\ 

        DTC & 128 & \texttt{Mean} & 4.5 & 1.0 & T & F & F & .58 & .42 & 64 & 1 & 16.6 & 3.42 & 0.87 \\ 

        \hline

        TC & 192 & \texttt{Sum} & 7.9 & 1.5 & T & F & T & .42 & .32 & 32 & 5 & \textbf{521} & \textbf{188} & \textbf{24.7} \\ 

        GH & 256 & \texttt{Sum} & 6.1 & 1.0 & T & F & T & .93 & .20 & 64 & 3 & 1992 & 340 & \textbf{32.1} \\ 
        
    \end{tabular}
    \caption{Best trials' hyperparameters for concurrent optimization of NFNs ($\eta \times 10^{-4}$).}
    \label{tab:doom_nfn_best_hyperparameters}
\end{table}
\vspace{-16pt}

\subsubsection{Attaining Weapon Competency} 




DNNs and NFNs competently handled weapons for their basic usage. 
For BR, variance equality was investigated by a F-test and a significantly large difference between DNN (SD=10.0) and NFN (SD=41.6) was present, $F(24, 24)$=$0.06$, $p$<.0001. 
Thus, a Welch's $t$-test measured the discrepancy in the BR scenario and found it to be statistically significant, $t(48)$=
$-2.4775$, 
$p$=$.0198
, 
d$=$-.7007$
. Equality of variances held for BP and PP, $F(24,24)$=$1$, $p$=$.963$ and $F(24,24)$=$0.9, p$=$.843$, respectively. 
Differences were statistically insignificant for BP and PP with $t(48)$=$0.4569, p $=$.6498$, $d $=$ 
.1292$ 
and $t(48) $=$ 0.5050, p $=$ .6159, d $=$ 
.1428$, respectively.

Due to its simplicity, 1.5-\texttt{entmax} could substitute \texttt{softmax} to produce sparse firing levels ($\alpha$=1.5) in the BP task. Additionally, delayed neurogenesis ($+\mu$>1), CFs (CF=T), and layer normalization (LN=T) helped the NFN in the BP scenario; interestingly, this is the only scenario where STE outperforms STGE, but STGE's best trial (not shown in Table~\ref{tab:doom_nfn_best_hyperparameters}) did achieve a Mean, SD, and $m$ of 83.12, 9.30, and 26.3, respectively. BR and PP used an unconstrained STGE ($\theta =0$), no layer normalization (LN=F), and retained the sampled Gumbel noise for a substantial amount of time ($\mathbf{N} \geq 128$). BR and PP required the same number of fuzzy logic rules ($|U|$=$256$), too. 

Of all three scenarios, the hyperparameters for PP should likely be given more attention, as BP and BR are more straightforward by comparison. 

\subsubsection{Successfully Defending Position}
DNN or NFN were both successful in holding the line (HTL) or defending the center (DTC) from enemy waves. 
Equality of variances can reasonably be assured for HTL but not for DTC, $F(24,24)$=$1.1, p$=$.902$ and $F(24,24)$=$0.3, p$=$.008$, respectively. 
There was no statistically significant difference in the HTL scenario, $t(48)$=$-1.1827, p$=$.2428, 
d$=$-.3345
$. Furthermore, the evaluation performance of DNNs and NFNs was comparable in the much more challenging DTC scenario, $t(48)$=$-1.6744, p$=$.1022, 
d$=$-.4736$.

Best trials for HTL and DTC used an unconstrained STGE ($\theta$=0) with Gumbel noise retained ($32 \leq \mathbf{N} \leq 64$), \CuiEtAl's Eq.~\ref{eq:mu_rule_mean} ($w_{u}$=\texttt{Mean}) and an approximately equal value for $\epsilon$ ($\sim$0.40). Disagreements occur in the remaining hyperparameters. In HTL, where all enemies are placed before the player, the agent could use $1.5$-\texttt{entmax}, CFs, twice the value of $\tau$, and delay neurogenesis ($+\mu$=3). In contrast, in DTC, where the agent must maintain situational awareness, the best performance was attained with \texttt{softmax} ($\alpha$=1) as all firing levels were necessary, stochastic behavior of STGE was less exploratory ($\tau$=$0.58$), and neurogenesis was immediate ($+\mu$=1).

Although HTL is challenging, it involves less situational awareness than DTC, where enemies may surround the player. Thus, the hyperparameters found to work best in DTC may be given greater priority over those found in HTL, as the configurations reported in Table~\ref{tab:doom_nfn_best_hyperparameters} are not the only trials that performed well. The overall configuration of the NFN for DTC is shared with PP ($w_{u}=\texttt{Mean}$ $\alpha=1.0$, STGE, no CFs, no LN, $\epsilon \approx 0.40$, $+\mu=1$).

\subsubsection{Achieving Proficient Evasion \& Navigation}
NFNs and DNNs successfully learned to take cover (TC) and gather health (GH). Most noticeably, the NFNs managed to maintain a slight advantage in TC, whereas DNNs mastered GH. There was a statistically significant difference in the equality of sample variances for TC and GH, $F(24,24)$=$2.5$, $p$=$.029$ and $F(24,24)$=$0, p$<.001, respectively. Though, the difference in evaluation performance was still not statistically significant, $t(48)$=$0.5676$, $ p$=$.5735$, $d$=$
.1605$ 
and $t(48)$=$1.5882$, $p$=$.1253, 
d$=$.4492$, respectively. 

The best trials of these scenarios for NFNs revealed a common configuration; NFNs in both TC and GH used an unconstrained STGE ($\theta$=0) with Gumbel noise retained ($32 \leq \mathbf{N} \leq 64$), Eq.~\ref{eq:mu_rule_sum} ($w_{u}$=\texttt{Sum}), layer normalization (LN=T), and delayed neurogenesis ($+\mu\geq3$). NFNs can also utilize sparse firing levels in TC with 1.5-\texttt{entmax}, while concurrent optimization of the NFNs adapts to each scenario through different values of $\tau$ and $\epsilon$. 

\subsubsection{Discussion of Optimal Hyperparameters}
The only conclusion that might be immediately drawn from Table~\ref{tab:doom_nfn_best_hyperparameters} is that STGE was almost always preferred to STE. Rather than resampling a new Gumbel noise for every forward pass, retaining the sampled Gumbel noise in STGE for some time was consistently beneficial ($\mathbf{N}$>1), as it allowed greater exploration of a possible mutation in the NFN's FRB. Resoundingly, it was found that the binary membership matrix, $\mathbf{M}$, should \textit{not} be constrained ($\theta$=0) according to scalar cardinality, $\mathbf{S}$, when using STGE. This may have occurred since the constraints were derived from the current batch information, which was likely fluctuating across different batches, inevitably leading to destabilization. Further investigation may be needed, but in its current form, it remains ineffective. Alternatively, this may suggest that STGE alone is sufficient in avoiding invalid or problematic FRBs with respect to (w.r.t.) $\epsilon$-completeness (refer to~\ref{appendix:constraint} for more).

As previously mentioned, the configurations for PP and DTC are identical regarding overall design choices for concurrent optimization. 
Scenarios PP and DTC are particularly challenging as they require estimating/anticipating enemy positions and maintaining situational awareness in the latter. Thus, the recurrence of this single design configuration in these challenging tasks suggests greater importance. Still, best trials for TC and GH also share a common design choice, except 1.5-\texttt{entmax} is used in TC. Neither of these two potential configurations integrates beliefs via CFs. 
Across all scenarios where leveraging 1.5-\texttt{entmax} was best (BP, HTL, TC), 
Tree Parzen estimator also found that CFs, layer normalization, or both to incorporate beliefs/amplify firing levels worked best. In contrast to the other four scenarios, where all configurations only used \texttt{softmax}, either CF or layer normalization was incorporated, but not both. This suggests that 1.5-\texttt{entmax} may require additional mitigation strategies for the firing levels due to its sparse output. Also, 1.5-\texttt{entmax} was typically only reserved for the more straightforward scenarios \textemdash{} BP, HTL, or TC, suggesting a significant reduction in the entire FRB's simultaneous firing (via $\alpha$=1.5) may be too aggressive for others that require more nuance. Future work could consider exploring adaptive $\alpha$-entmax with $\alpha$ values that may range within $(1.0, 1.5)$. 

However, mitigation strategies were also more prevalent when \CuiEtAl's Eq.~\ref{eq:mu_rule_mean} did not substitute Eq.~\ref{eq:mu_rule_sum}. In particular, whenever preliminary firing level calculation was unchanged (i.e., Eq.~\ref{eq:mu_rule_sum}), some form of mitigation was incorporated \textemdash{} CF (BP, BR) or layer normalization (BP, TC, GH). Instead, \CuiEtAl's Eq.~\ref{eq:mu_rule_mean} required additional mitigation in just one scenario, HTL. 
Every configuration relying on \CuiEtAl's Eq.~\ref{eq:mu_rule_mean} involved unconstrained STGE with Gumbel noise retained, no layer normalization, similar values for $\epsilon$, and performed immediate or delayed neurogenesis depending on if it used \texttt{softmax} or 1.5-\texttt{entmax}, respectively. All configurations with 1.5-\texttt{entmax} were paired with delayed neurogenesis, and every configuration that practiced immediate neurogenesis never used layer normalization or 1.5-\texttt{entmax}.

\subsubsection{Synthesis of Optimal Hyperparameters}
We sought to synthesize each design choice's impact on NFN performance to establish a consensus on key factors for effective concurrent optimization. Hyperparameter importance may be calculated with different techniques, but three were explored here: (1) Functional ANOVA (fANOVA) \cite{pmlr-v32-hutter14}, (2) Mean Decrease Impurity (MDI) \cite{10.5555/2999611.2999660}, and (3) PED-ANOVA (10\% Quartile) \cite{10.24963/ijcai.2023/488}. 

\begin{table}
\centering
\footnotesize
\setlength\extrarowheight{-3pt}
\addtolength{\tabcolsep}{-0.175em}
\caption{Hyperparameter importance for concurrently optimizing NFNs in DOOM tasks.}
\label{tab:hyperparameter_importance}
\begin{tabular}{c||c||RRRRRRRRRRRR}
\toprule
Task & Evaluator & \multicolumn{1}{c}{STGE} & \multicolumn{1}{c}{$\theta$} & \multicolumn{1}{c}{LN} & \multicolumn{1}{c}{$w_{u}$} & \multicolumn{1}{c}{$\tau$} & \multicolumn{1}{c}{$\eta$} & \multicolumn{1}{c}{$\epsilon$} & \multicolumn{1}{c}{$\alpha$} & \multicolumn{1}{c}{$\mathbf{N}$} & \multicolumn{1}{c}{$|U|$} & \multicolumn{1}{c}{$+\mu$} & \multicolumn{1}{c}{CF} \\
\hline\hline
 & fANOVA & .41 & .36 & .04 & .03 & .03 & .03 & .03 & .03 & .02 & .01 & .00 & .00 \\
BP & MDI & .32 & .16 & .04 & .05 & .09 & .08 & .07 & .05 & .06 & .03 & .02 & .01 \\
 & PED & .00 & .05 & .00 & .00 & .17 & .05 & .29 & .08 & .11 & .24 & .00 & .00 \\
\hline

 & fANOVA & .33 & .49 & .03 & .03 & .01 & .03 & .02 & .05 & .01 & .00 & .00 & .00 \\
BR & MDI & .28 & .28 & .08 & .07 & .03 & .06 & .05 & .08 & .03 & .02 & .01 & .01 \\
 & PED & .16 & .35 & .02 & .01 & .13 & .06 & .06 & .08 & .01 & .10 & .01 & .01 \\
\hline

 & fANOVA & .36 & .39 & .04 & .04 & .02 & .06 & .02 & .05 & .02 & .01 & .00 & .00 \\
PP & MDI & .28 & .09 & .08 & .08 & .08 & .11 & .08 & .07 & .05 & .03 & .03 & .02 \\
 & PED & .01 & .04 & .00 & .03 & .23 & .31 & .06 & .15 & .04 & .02 & .12 & .00 \\
\hline

 & fANOVA & .41 & .43 & .02 & .06 & .01 & .02 & .01 & .03 & .01 & .00 & .00 & .00 \\
HTL & MDI & .20 & .20 & .10 & .10 & .06 & .08 & .06 & .11 & .05 & .02 & .01 & .01 \\
 & PED & .07 & .13 & .01 & .00 & .07 & .41 & .10 & .00 & .09 & .01 & .03 & .08 \\
\hline

 & fANOVA & .34 & .38 & .03 & .04 & .03 & .04 & .04 & .05 & .02 & .01 & .01 & .00 \\
DTC & MDI & .18 & .12 & .05 & .05 & .12 & .13 & .12 & .05 & .07 & .05 & .03 & .02 \\
 & PED & .03 & .10 & .01 & .07 & .15 & .21 & .18 & .04 & .03 & .08 & .08 & .01 \\
\hline

 & fANOVA & .45 & .30 & .06 & .05 & .01 & .04 & .02 & .03 & .02 & .00 & .01 & .00 \\
TC & MDI & .17 & .23 & .08 & .09 & .07 & .08 & .06 & .06 & .06 & .03 & .03 & .02 \\
 & PED & .01 & .13 & .06 & .03 & .26 & .15 & .06 & .00 & .10 & .16 & .01 & .00 \\
\hline

 & fANOVA & .35 & .32 & .01 & .01 & .03 & .11 & .04 & .04 & .06 & .01 & .01 & .01 \\
GH & MDI & .21 & .20 & .04 & .02 & .10 & .13 & .09 & .05 & .08 & .03 & .03 & .02 \\
 & PED & .04 & .40 & .01 & .08 & .14 & .07 & .06 & .01 & .05 & .07 & .06 & .00 \\
\bottomrule
\end{tabular}
\end{table}

\begin{table}[ht]
    \centering
    \addtolength{\tabcolsep}{-0.3em}  
    \begin{tabular}{c||c||c|c|c}
        Task & $W$ & $F(11, 24)$ & ICC$(3,k)$ & 95\% CI \\
        \hline\hline

        BP & .5804 ($p=.0584$) & 1.6533 ($p=.1464$) & .1788 & (-0.14, 0.59) \\
        BR & .7545 ($p=\mathbf{.0094}$) & 15.6992 ($p\leq \mathbf{.0001}$) & \textbf{.8305} & (0.63, 0.94) \\
        PP & .5618 ($p=.0699$) & 1.4950 ($p=.1975$) & .1416 & (-0.16, 0.56) \\        

        \hline

        HTL & .5260 ($p=.0977$) & 2.1521 ($p=.0565$) & .2775 & (-0.06, 0.66) \\
        DTC & .6659 ($p=\mathbf{.0246}$) & 2.0473 ($p=.0690$) & .2588 & (-0.07, 0.65) \\        

        \hline

        TC & .5618 ($p=.0699$) & 1.9325 ($p=.0859$) & .2371 & (-0.09, 0.63) \\
        GH & .6550 ($p=\mathbf{.0275}$) & 6.2592 ($p=\mathbf{.0001}$) & \textbf{.6368} & (0.32, 0.86) \\        
        
    \end{tabular}
    \caption{Hyperparameters' rankings and importance for concurrent optimization of NFNs.}
    \label{tab:hyperparameter_analysis}
\end{table}

Hyperparameters' importance varied between these methods and across the different tasks \textemdash{} BP, BR, PP, DTC, HTL, TC, and GH (Table~\ref{tab:hyperparameter_importance}). 
Each hyperparameter was ranked based on its importance w.r.t. every evaluation technique. A non-parametric statistical test called the Kendall's coefficient of concordance (i.e., Kendall's $W$) was then conducted to determine if there was a statistically significant rank correlation per task. Additionally, the absolute reliability of the three feature importance methods was assessed using the Intraclass Correlation Coefficient, ICC(3,$k$), per task, as well. Results are located in Table~\ref{tab:hyperparameter_analysis}. 
Agreement w.r.t. the hyperparameters' rankings appears rather substantial across each task according to Kendall's $W$, but only a few of these agreements were statistically significant w.r.t. $p$-values (e.g., BR, DTC, GH). As for absolute reliability of the three feature importance methods according to ICC(3,$k$), only BR and GH had statistically significant excellent (ICC(3, $k$)=.8305, $p \leq .0001$) and good (ICC(3, $k$)=.6368, $p$=$.0001$) ICC inter-rater agreements \cite{Cicchetti1994GuidelinesCA}, respectively; inter-rater agreement on other tasks remained poor (i.e., ICC(3, $k$) < .40), though, there was some marginal statistical significance on tasks such as HTL, DTC, TC.

Although there was statistical agreement on hyperparameters' ranking and reliability in tasks BR and GH, the presence of poor ICC inter-rater agreements across most tasks implies hyperparameters' importance measured by fANOVA, MDI, and PED-ANOVA largely remain inconsistent. This can be visually validated by consulting Table~\ref{tab:hyperparameter_analysis} for either BR or GH; for instance, fANOVA calculated STGE had an importance of .33 in the BR scenario, whereas the value of .16 is assigned by PED-ANOVA, despite ICC(3,$k$) suggesting excellent inter-rater agreement. 
This indicates that the parameters that drive global variance (fANOVA) are entirely distinct from those that drive the surrogate model's decisions (MDI) and those that drive optimal performance (PED-ANOVA). Therefore, we cannot create a reliable consensus on hyperparameter importance between all of these evaluators, but must select one to emphasize based on our research objective. Given the demonstrated independence, we selected PED-ANOVA as the primary evaluator due to its practical utility in directly isolating sensitive hyperparameters within the high-performance region of the search space (e.g., fANOVA may be biased toward unimportant subspaces \cite{10.24963/ijcai.2023/488}).

In an effort to aggregate the results to obtain a more reliable conclusion regarding which hyperparameters were important, but also, what values should be assigned to them, a clustering procedure was performed on all trials. Specifically, trials were considered similar based on a weighted Gower distance \cite{56fe9d33-d905-3182-860c-ed85b7b00402}, where the assigned weights were equal to the hyperparameters' corresponding PED-ANOVA importance values; the Gower distance was chosen due to the mixed-type data objects involved. Then, ordering points to identify the clustering structure (OPTICS) \cite{10.1145/304182.304187} was conducted. For each identified cluster, their pooled mean and pooled standard deviation were calculated from the trials that belonged to it. Clusters' medoids were determined by minimizing sum of distances. The medoid with the maximum pooled mean, for each task, are ultimately reported in Table~\ref{tab:ped_anova_nfn_best_hyperparameters}, where the design choices that explained 90\% of the performance gains are highlighted; identical design choices between Tables~\ref{tab:doom_nfn_best_hyperparameters} and~\ref{tab:ped_anova_nfn_best_hyperparameters} are reflected by bold font.

\begin{table}[ht]
    \centering
    \addtolength{\tabcolsep}{-0.3em}  
    \begin{tabular}{c||c|c|c|c|c|c|c|c|c|c|c|c||c|c}
        Task & $|U|$ & $w_{u}$ & $\eta$ & $\alpha$ & STGE & CF & LN & $\tau$ & $\epsilon$ &$\mathbf{N}$ & $+\mu$ & Mean & SD & Sup. \\
        \hline\hline

        BP & \hl{64} & \ttfamily{\fontseries{b}\selectfont Sum} & 9.9 & 1.0 & T & F & F & \hl{.92} & \hl{.32} & \hl{128} & 5 & 79.5 & 13.5 & 3 \\
        
        BR & \hl{\textbf{256}} & \ttfamily{\fontseries{b}\selectfont Sum} & \textbf{9.2} & \hl{\textbf{1.0}} & \hl{\textbf{T}} & \textbf{T} & \textbf{F} & \hl{\textbf{.42}} & \hl{\textbf{.35}} & \textbf{256} & \textbf{1} & -17.8 & 136.1 & 3 \\  

        PP & 128 & \ttfamily{\fontseries{b}\selectfont Mean} & \hl{6.8} & \hl{1.5} & \textbf{T} & \textbf{F} & \textbf{F} & \hl{.86} & \hl{.27} & 1 & \hl{5} & .14 & .42 & 4 \\ 

        \hline

        HTL & 256 & \texttt{Sum} & \hl{7.2} & 1.0 & F & \hl{\textbf{T}} & \textbf{F} & \hl{.74} & \hl{.36} & 128 & 5 & 21.04 & 5.59 & 3 \\ 

        DTC & \hl{\textbf{128}} & \hl{\ttfamily{\fontseries{b}\selectfont Mean}} & \hl{\textbf{4.5}} & \hl{\textbf{1.0}} & \textbf{T} & \textbf{F} & \textbf{F} & \hl{\textbf{.58}} & \hl{\textbf{.42}} & \textbf{64} & \hl{\textbf{1}} & 10.2 & 5.22 & 5 \\ 

        \hline

        TC & \hl{128} & \ttfamily{\fontseries{b}\selectfont Sum} & \hl{6.9} & \textbf{1.5} & \textbf{T} & \textbf{F} & \hl{\textbf{T}} & \hl{1.0} & .41 & \hl{\textbf{32}} & 1 & 448 & 252 & 4 \\ 

        GH & \hl{128} & \hl{\texttt{Mean}} & \hl{9.0} & \textbf{1.0} & \textbf{T} & \textbf{F} & F & \hl{.50} & \hl{.26} & \textbf{64} & \hl{5} & 1533 & 721 & 3 \\ 
        
    \end{tabular}
    \caption{Medoids of best hyperparameters for concurrent optimization of NFNs ($\eta \times 10^{-4}$).}
    \label{tab:ped_anova_nfn_best_hyperparameters}
\end{table}

Agreement of hyperparameters' rankings with respect to PED-ANOVA across all tasks revealed a statistically significant weak consensus, $W$=.4316, $p$=.0005. However, the absolute reliability of PED-ANOVA's hyperparameter importance across all tasks was excellent, ICC($3, k$)=.7546, 95\% CI (0.46, 0.92), $F(11, 66)$=$4.075078, p$=$.0001$.

\begin{figure}
    \centering
    \includegraphics[width=0.5\linewidth,trim={1.5cm 3.75cm 2.85cm 1.5cm},clip]{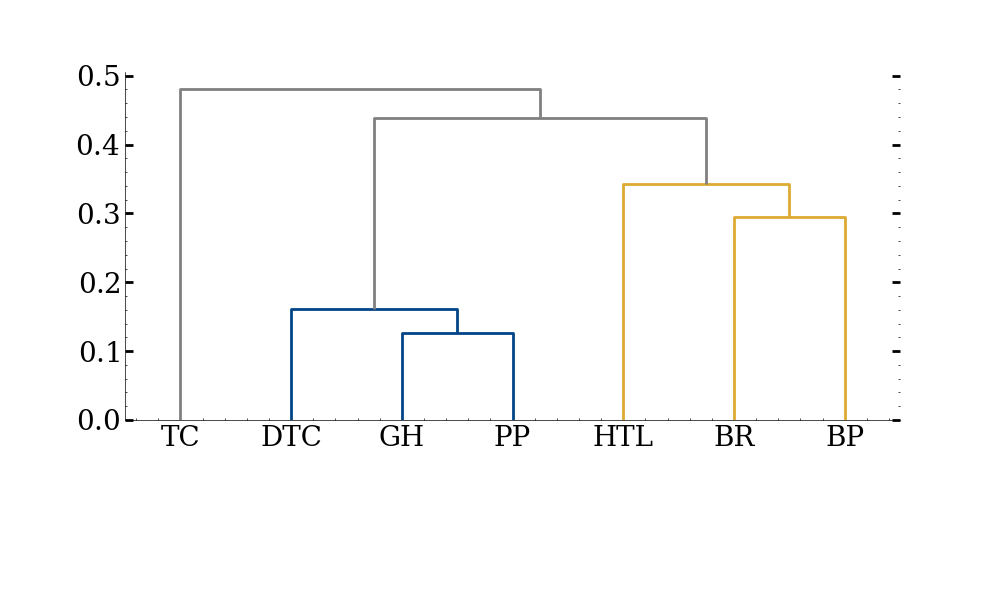}
    \caption{Agglomerative clustering of tasks with similar hyperparameter values.}
    \label{fig:dendrogram}
\end{figure}

Recurring themes, or patterns, on which hyperparameter values perform best across tasks may be directly observed from Table~\ref{tab:ped_anova_nfn_best_hyperparameters}, but we sought to more formally quantify or interpret these findings in a less subjective manner. Similarity between tasks' best hyperparameter medoids was determined with a weighted Gower distance once again; considering the absolute reliability of PED-ANOVA's hyperparameter importance across all tasks was significantly excellent, ICC($3, k$)=.7546 ($p$=.0001), the mean was calculated and utilized to weigh the Gower distance. Figure~\ref{fig:dendrogram} shows a dendrogram with average linkage via Unweighted Pair-Group Method using Arithmetic averages (UPGMA) where tasks with similar best hyperparameter medoids were grouped based on the calculated weighted Gower distance matrix. Therefore, these groupings not only account for how similar the hyperparameter values are, but also, are weighed by how important the hyperparameters were in contributing to optimal task performance. 

Figure~\ref{fig:dendrogram} reaffirms Table~\ref{tab:ped_anova_nfn_best_hyperparameters} by showing DTC, GH, and PP share similar values for important hyperparameters: rule count ($|U|$), premise aggregation ($w_{u}$), learning rate ($\eta$), Gumbel temperature ($\tau$), and minimum membership degree ($\epsilon$); additionally, these tasks also agreed on STGE, no Certainty Factors (CFs), and no Layer Normalization (LN), but not on neurogenesis ($+\mu$) since immediate neurogenesis ($+\mu$=1) performed best for DTC. Although, we still could possibly consider this a potentially viable configuration of hyperparameters that are transferable or generalizable between these tasks. 

Figure~\ref{fig:dendrogram} suggests HTL, BR, and BP share some similar values for important hyperparameters, but its grouping is less cohesive since its minimum and maximum values calculated by UPGMA were 
.2945 and .3426, 
respectively. These tasks had fewer similar hyperparameters as expected: $w_{u}$, $\alpha$, LN, $\epsilon$. 

Since the Gower distance was weighed by the mean PED-ANOVA's calculated hyperparameter importance, some hyperparameters were given greater emphasis in influencing the similarity than others, as they are more important with respect to optimal performance. Thus, disagreements for those that have lesser influence is not as concerning as it may appear. For instance, the following hyperparameters remained slightly contested across the tasks: $|U|$ ($=64$ for BP, mean importance =
.0966), STGE (=F for HTL, mean importance =
.0467), CF (=F for BP, mean importance =
.0145), $\mathbf{N}$ (=256 for BR, mean importance =
.0622), and $+\mu$ ($=1$ for BR, mean importance =
.0449), accounting for approximately 
26.49\%
of total mean importance. Thus, this too, could also be viewed as a potentially viable configuration of hyperparameters that are transferable or generalizable between these tasks.

Hyperparameters associated with performing well in TC are essentially a standalone configuration as it merges at a UPGMA value of 
.4813, after the prior two clusters merge 
(at a UPGMA of .4392). This suggests that the design choices required to take cover from, or evade, enemy attacks when the ability to defend oneself is prohibited, are substantially different from tasks where the player's agent might be capable of attacking (e.g., HTL, BR, BP). 

Tasks HTL, BR, and BP share extremely alike rooms and restrict the player's movement. The player is free to attack in all scenarios, but in HTL, there are never-ending spawns of enemy monsters attacking the player, whereas this does not occur in BR or BP. Thus, this commonality could be why they shared similar hyperparameters. Tasks DTC, GH, and PP require more situational awareness and the capability to follow targets or predict positions. For example, this could be predicting positions of an enemy (DTC \& PP), or where a medkit might land when it falls from the sky (GH). Additionally, the player agent must maintain situational awareness in all of these scenarios; either by searching for and tracking flanking enemies (DTC \& PP), or finding medkits outside of the player's current field of view (GH).

Every proposed design option (e.g., $\alpha$, STGE, LN) in this article (with the exception of the $\theta$ constraint) was found to be helpful and important in at least one DOOM task. However, ultimately, tasks DTC, PP and GH (as well as TC) are substantially more difficult than tasks HTL, BR, and BP. Greater emphasis should thus instead be placed on the findings for tasks DTC, PP, and GH (as well as TC). Notably, \CuiEtAl's suggestion of using the $\texttt{Mean}$ instead of \texttt{Sum}, STGE for our proposed technique of GBNA, retaining or sampling Gumbel noise $\mathbf{N}$, immediate or delayed neurogenesis ($+\mu$), were all important factors when succeeding in these experimental tasks. 

Future work will continue this research to investigate whether a particular hyperparameter configuration can be discovered that performs universally well across tasks. In the meantime, these results remain extremely encouraging; it is imperative to gently remind the reader that concurrently optimizing the NFNs' structure and parameters with gradient-based neuroplastic adaptation (GBNA) works without requiring any modifications to the learning algorithms, assumptions, etc. The hallmark accomplishment of this article is the capability of adapting NFNs' structure and parameters in a general, application-independent approach, that can even work in online, complex, vision-based RL. Although not demonstrated here, as this is a special issue for interpretable RL, this methodology for adapting NFNs would also work well in other paradigms, such as supervised or unsupervised learning. 


\section{Discussion}
NFNs blend fuzzy systems' transparency with neural architectures' adaptability. 
Concurrent optimization of NFNs 
is a significant advancement in allowing NFNs to seamlessly integrate with various neural architectures by not requiring a predefined architecture. We investigated the trade-offs between using STE or STGE for generating compound condition attributes of fuzzy logic rules and found that STGE was oftentimes crucial for the concurrent optimization of NFNs as it consistently outperformed STE. Therefore, GBNA should usually be implemented with STGE, as the stochastic sampling and exploration of alternatives proved incredibly effective; in only one medoid of best hyperparameters was STE selected instead (HTL), but it was also not considered a significant contributor to the task performance (7\%) w.r.t. PED-ANOVA. We also verified \CuiEtAl's suggestion of Eq.~\ref{eq:mu_rule_mean} is particularly effective in challenging high-dimensional vision-based tasks (PP, DTC, GH). Additionally, GBNA is reassuringly robust to immediate neurogenesis (BR, DTC, TC), suggesting it is capable of dynamically abiding by $\epsilon$-completeness, but creating new fuzzy sets can be delayed, too (BP, PP, HTL, GH); possible task state complexity appears to influence whether neurogenesis should be immediate or delayed, as DTC and TC are much more dynamic scenarios than BP, PP, GH, or even HTL. Finally, other design choices (e.g., incorporating beliefs into NFN's fuzzy logic rules) may also enhance performance in certain scenarios without hampering performance in other tasks. Numerical hyperparameters, such as fuzzy logic rule count, appear to remain task-dependent, similar to DNNs' hidden neuron count.

The holistic integration of GBNA with immediate, or delayed, neurogenesis constitutes the overall paradigm for the concurrent optimization of NFNs, and was a resounding success in training NFNs to play an FPS video game called DOOM. This opens up new possibilities for NFN research by 
eliminating exhaustive searches via genetic learning or unsupervised algorithms to find ideal neuro-fuzzy architectures. Unlike conventional NFNs committed to a fixed architecture, NFNs with concurrent optimization can dynamically redesign their parameters and structure based on a loss with respect to an objective measure. \textit{This generality of our NFN allows it to be the first capable of a variety of work, such as online/offline RL, vision-based tasks, and more.}

\section*{Acknowledgments}
NSF Grants supported this research: Integrated Data-driven Technologies for Individualized Instruction in STEM Learning Environments (1726550), CAREER: Improving Adaptive Decision Making in Interactive Learning Environments (1651909), and Generalizing Data-Driven Technologies to Improve Individualized STEM Instruction by Intelligent Tutors (2013502).

\appendix



\section{Constraining Straight-Through Gumbel Estimator}\label{appendix:constraint}

We developed a \textit{constrained STGE} to differentiably sample premises for fuzzy logic rules in an attempt to satisfy $\epsilon$-completeness and avoid numerical underflow. The constrained STGE only selects premises with strong and \textit{frequent} activation to avoid invalid selections of premises for fuzzy logic rules based on \textit{scalar cardinality}, $\mathbf{S}$ \cite{chen_mining_2011}. 
A threshold is found from \textbf{S} by relying on the $\theta^{\text{th}}$ percentile of each condition attribute, such that any term with $\mathbf{S}_{i, j} < \theta^{\text{th}}$ is temporarily removed, $\mathbf{M}' = (\mathbf{M} \odot \mathbf{S}) > \theta$ from consideration by the constrained STGE. 
As $\theta$ is decreased, the STGE becomes less constrained, and vice versa. We ultimately found constraining STGE was harmful in our experiments, but still include it in this paper for research transparency. 

\section{Amplifying Activation of Fuzzy Logic Rules}\label{appendix:firing_levels}
\begin{figure*}[ht]
    \centering
    \hspace{0.5in}\includegraphics[width=0.2\linewidth]{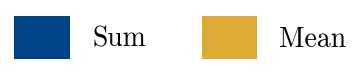} \\
    \vspace{-0.15in}
    \begin{tabular}{ccc}
        \multirow{7}{*}{\textit{(a)}} & \textit{(1) Unsorted} & \textit{(2) Sorted} \\
        \multirow{7}{*}{\textit{(b)}} & \includegraphics[width = 0.4\linewidth]{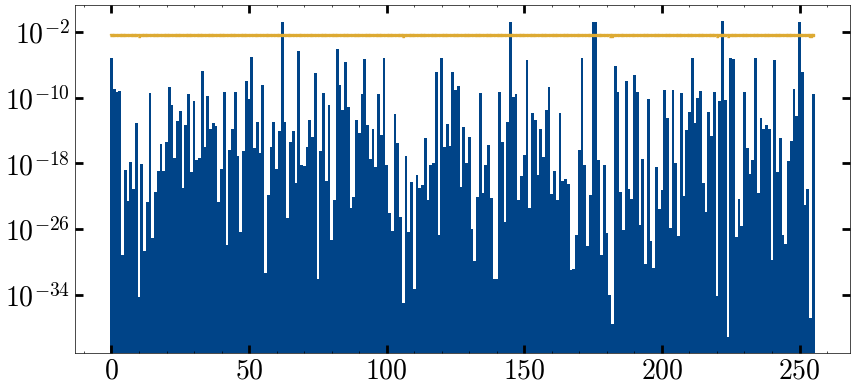} &
        \includegraphics[width = 0.4\linewidth]{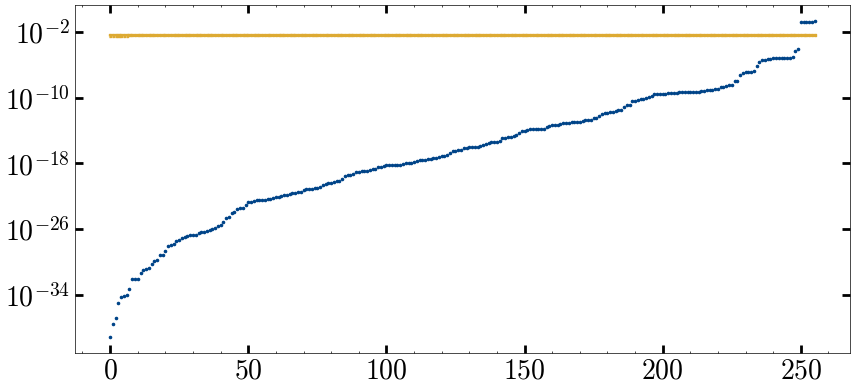} \\
        \multirow{7}{*}{\textit{(c)}} &\includegraphics[width = 0.4\linewidth]{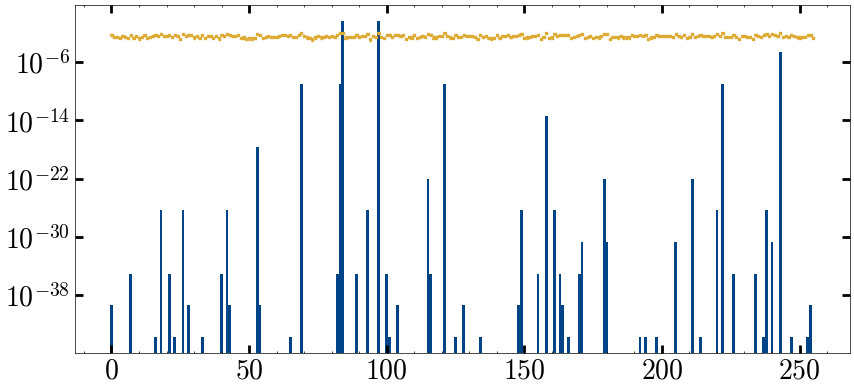} &
        \includegraphics[width = 0.4\linewidth]{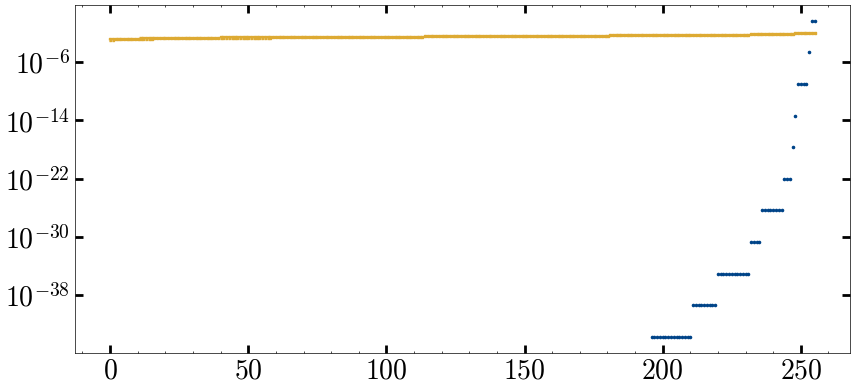} \\
        & \includegraphics[width = 0.4\linewidth]{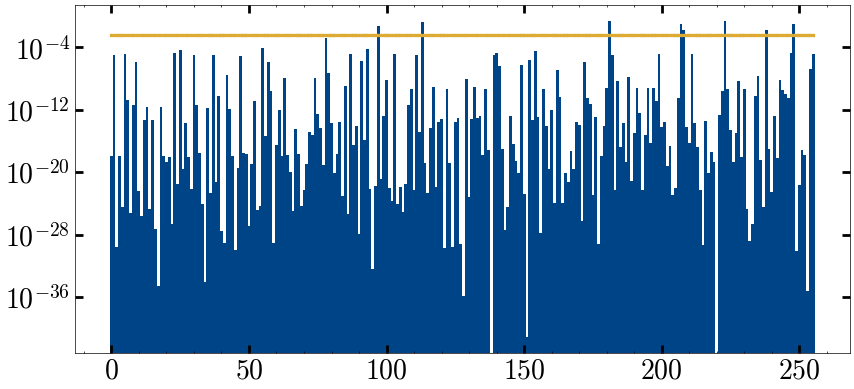} &
        \includegraphics[width = 0.4\linewidth]{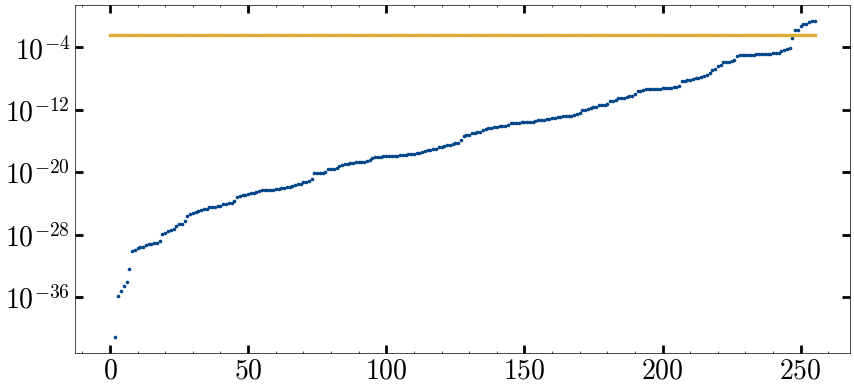} \\
    \end{tabular}
    \caption{Example responses of fuzzy logic rules with no amplification.}
    \label{fig:rule_cod_no_layer_norm_softmax}
\end{figure*}

\begin{figure*}[ht]
    \centering
    \hspace{0.5in}\includegraphics[width=0.2\linewidth]{appendix/figures/FiringLevels/legend.png} \\
    \vspace{-0.15in}
    \begin{tabular}{ccc}
        \multirow{7}{*}{\textit{(a)}} & \textit{(1) Unsorted} & \textit{(2) Sorted} \\
        \multirow{7}{*}{\textit{(b)}} & \includegraphics[width = 0.4\linewidth]{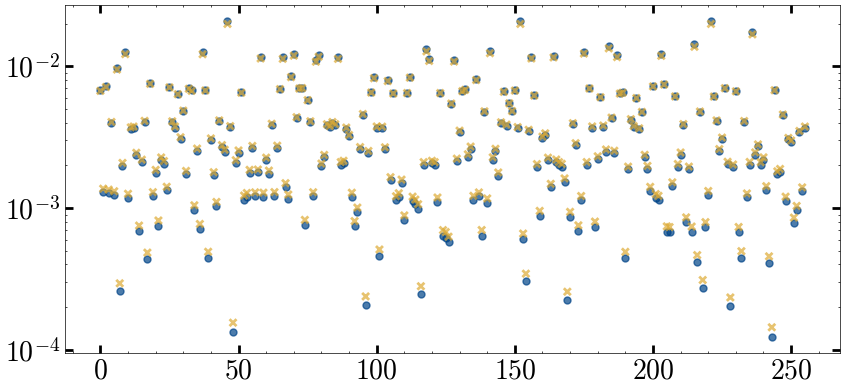} &
        \includegraphics[width = 0.4\linewidth]{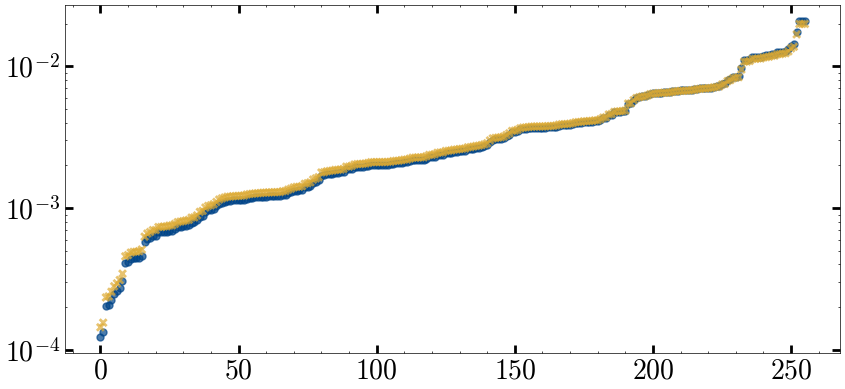} \\
        \multirow{7}{*}{\textit{(c)}} &\includegraphics[width = 0.4\linewidth]{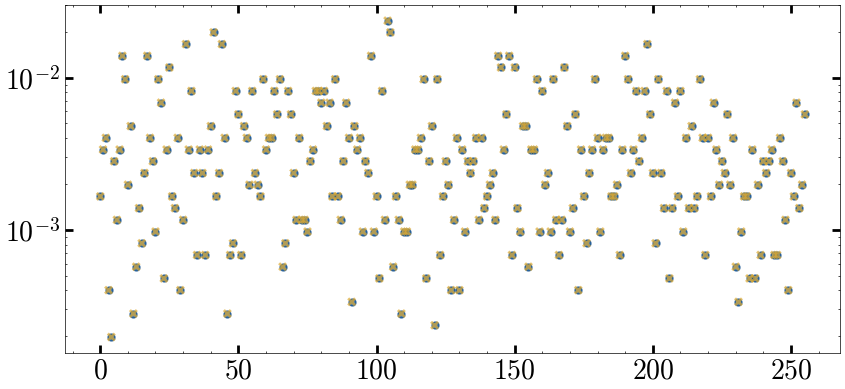} &
        \includegraphics[width = 0.4\linewidth]{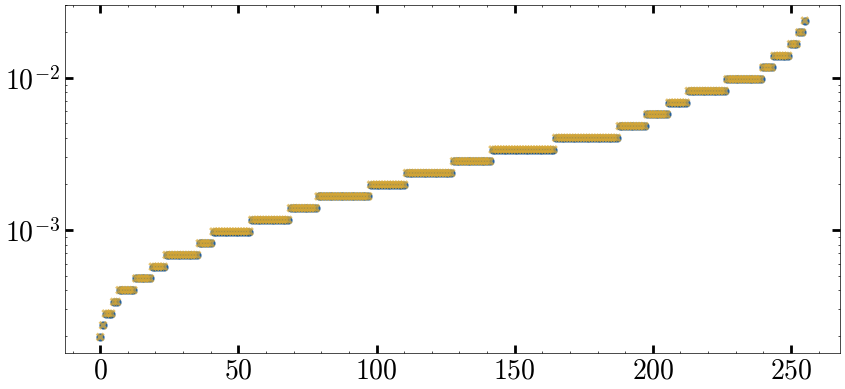} \\
        & \includegraphics[width = 0.4\linewidth]{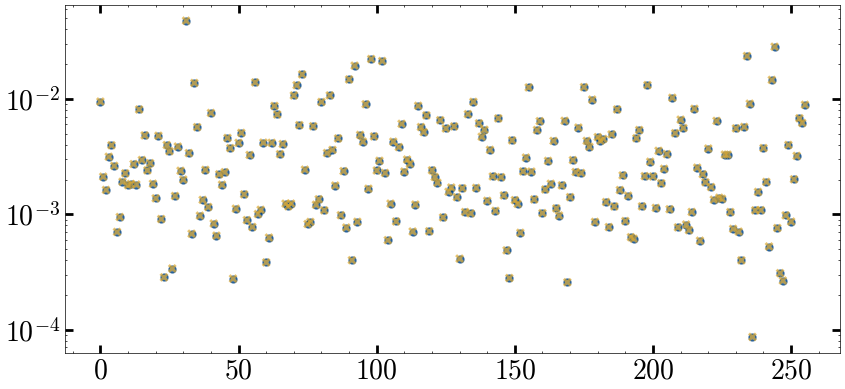} &
        \includegraphics[width = 0.4\linewidth]{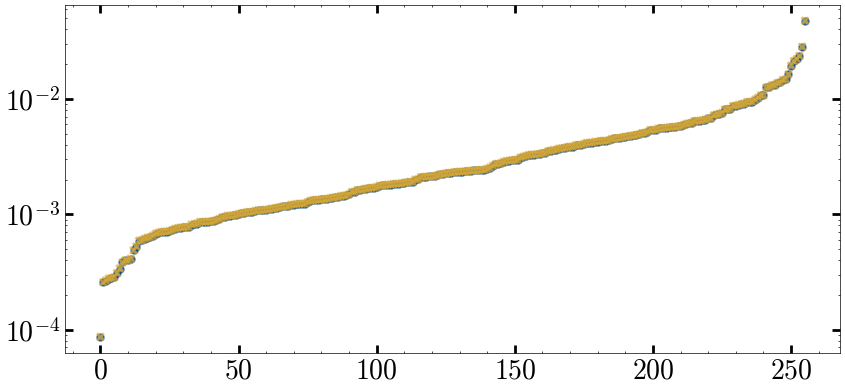} \\
    \end{tabular}
    \caption{Firing levels are amplified if layer normalization is incorporated.}
    \label{fig:rule_cod_layer_norm_softmax}
\end{figure*}

\begin{figure*}[ht]
    \centering
    \hspace{0.5in}\includegraphics[width=0.2\linewidth]{appendix/figures/FiringLevels/legend.png} \\
    \vspace{-0.15in}
    \begin{tabular}{ccc}
        \multirow{7}{*}{\textit{(a)}} & \textit{(1) Unsorted} & \textit{(2) Sorted} \\
        \multirow{7}{*}{\textit{(b)}} & \includegraphics[width = 0.4\linewidth]{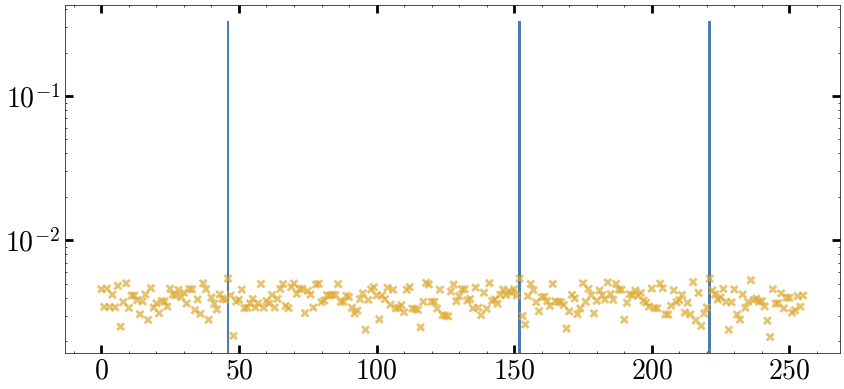} &
        \includegraphics[width = 0.4\linewidth]{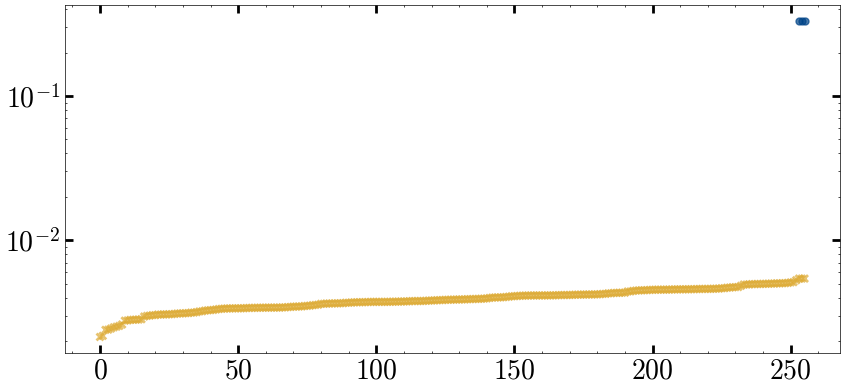} \\
        \multirow{7}{*}{\textit{(c)}} &\includegraphics[width = 0.4\linewidth]{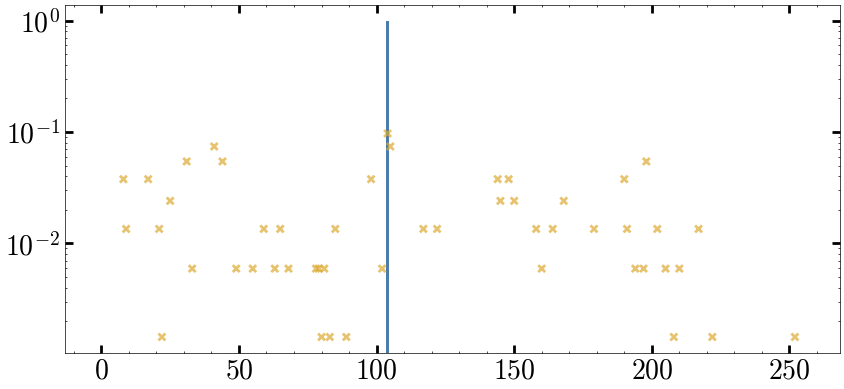} &
        \includegraphics[width = 0.4\linewidth]{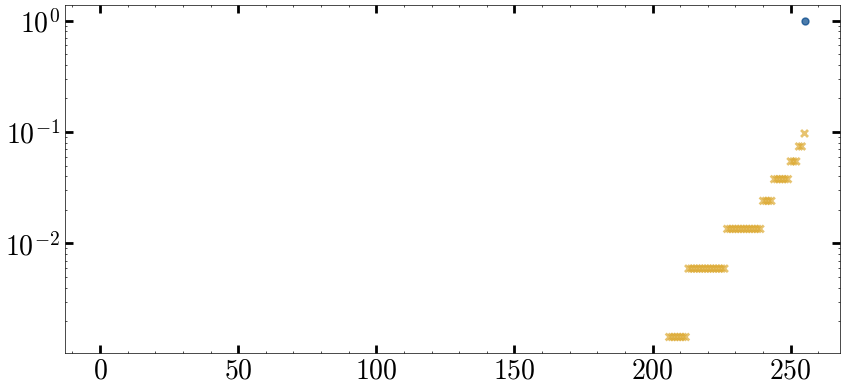} \\
        & \includegraphics[width = 0.4\linewidth]{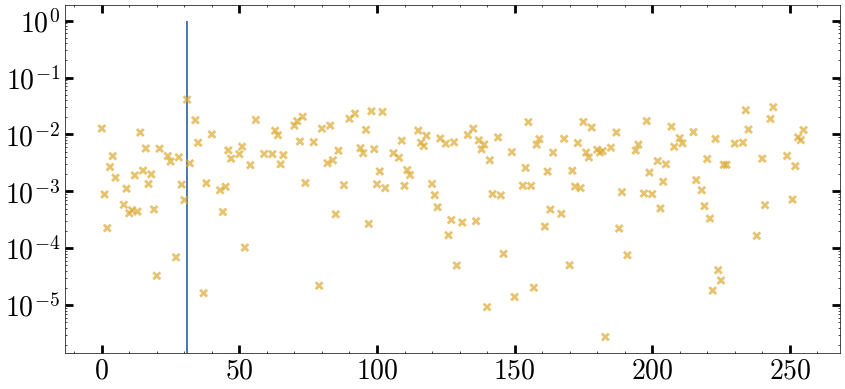} &
        \includegraphics[width = 0.4\linewidth]{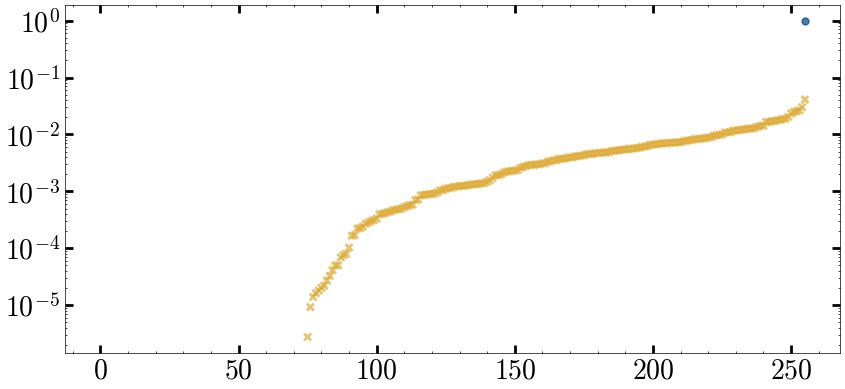} \\
    \end{tabular}
    
    \caption{Sparse firing levels occur if 1.5-\texttt{entmax} is used instead of \texttt{softmax}.}
    \label{fig:rule_cod_no_layer_norm_entmax}
\end{figure*}

\begin{figure*}[ht]
    \centering
    \hspace{0.5in}\includegraphics[width=0.2\linewidth]{appendix/figures/FiringLevels/legend.png} \\
    \vspace{-0.15in}
    \begin{tabular}{ccc}
        \multirow{7}{*}{\textit{(a)}} & \textit{(1) Unsorted} & \textit{(2) Sorted} \\
        \multirow{7}{*}{\textit{(b)}} & \includegraphics[width = 0.4\linewidth]{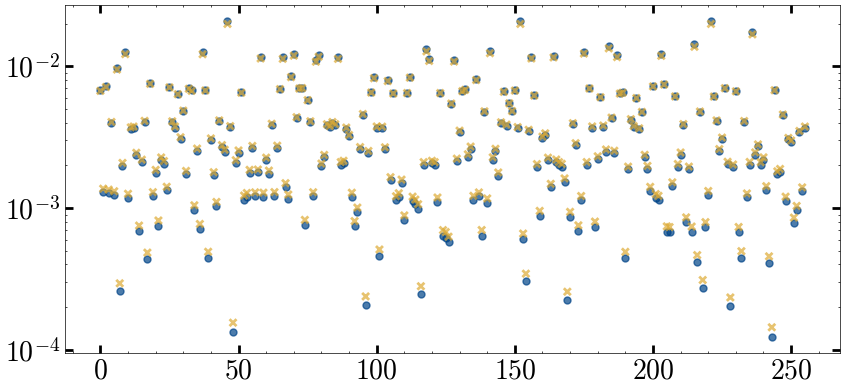} &
        \includegraphics[width = 0.4\linewidth]{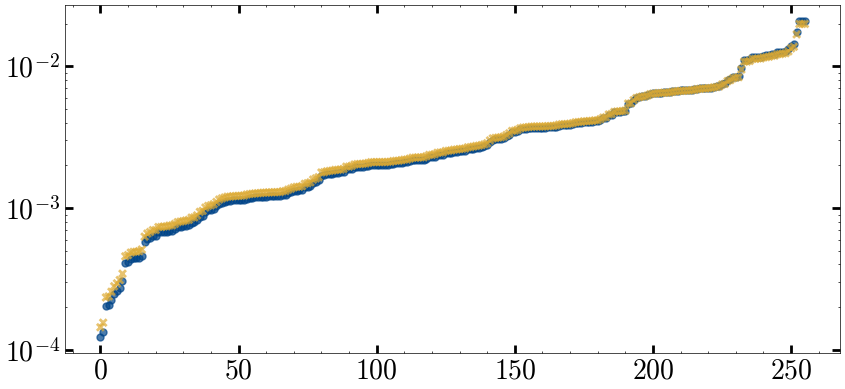} \\
        \multirow{7}{*}{\textit{(c)}} &\includegraphics[width = 0.4\linewidth]{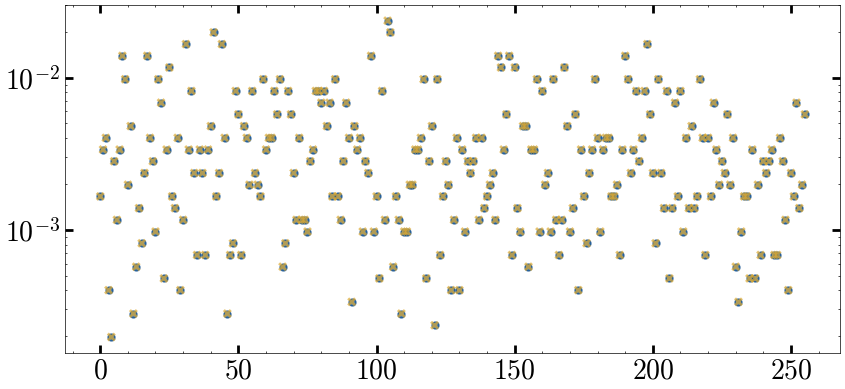} &
        \includegraphics[width = 0.4\linewidth]{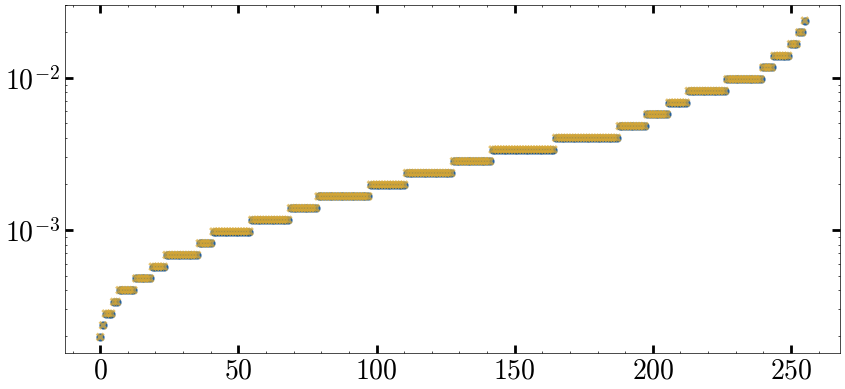} \\
        & \includegraphics[width = 0.4\linewidth]{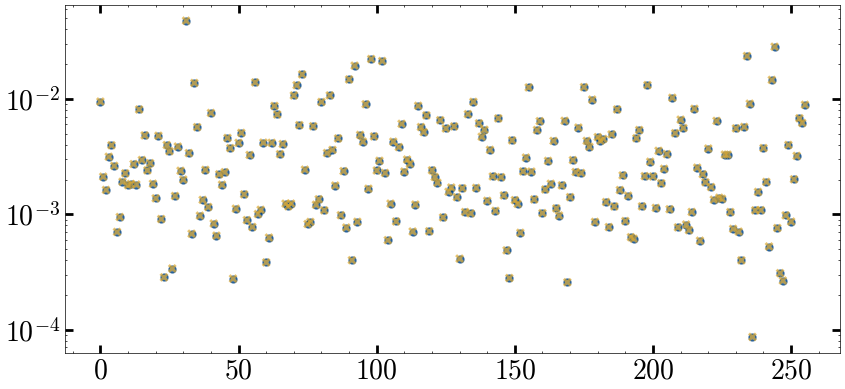} &
        \includegraphics[width = 0.4\linewidth]{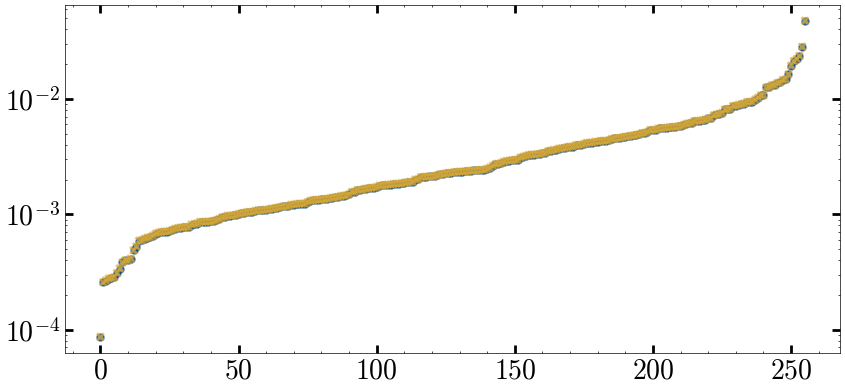} \\
    \end{tabular}
    \caption{Layer normalization with 1.5-\texttt{entmax} removes sparse firing levels.}
    \label{fig:rule_cod_layer_norm_entmax}
\end{figure*}

An FRB is randomly initialized with 256 fuzzy logic rules, each with 1600 condition attributes. For three randomly sampled observations (\textit{a}, \textit{b}, and \textit{c}) from a scenario in the DOOM video game, their corresponding activations, or firing levels, are illustrated in this section to showcase how dimensionality, with respect to condition size, as well as the number of fuzzy logic rules, may negatively hamper the firing levels. For each figure in this section, the x-axis indexes an individual rule, and the y-axis corresponds to its associated firing level (log scale). Activations determined by Eq.~\ref{eq:mu_rule_sum} are in blue, whereas those calculated by \CuiEtAl's Eq.~\ref{eq:mu_rule_mean} are shown in orange. Figures~\ref{fig:rule_cod_no_layer_norm_softmax}, \ref{fig:rule_cod_layer_norm_softmax}, \ref{fig:rule_cod_no_layer_norm_entmax} and \ref{fig:rule_cod_layer_norm_entmax} will all use the same three randomly sampled observations in their visualizations to facilitate direct comparisons between the firing levels. 

Figure~\ref{fig:rule_cod_no_layer_norm_softmax} shows normal inference (Eq.~\ref{eq:mu_rule_sum}) with an NFN against \CuiEtAl's Eq.~\ref{eq:mu_rule_mean}. Noticeably, Eq.~\ref{eq:mu_rule_sum} struggles to sufficiently fire fuzzy logic rules whereas \CuiEtAl's Eq.~\ref{eq:mu_rule_mean} greatly enhances activation levels, but uniformly applies this enhancement. Thus, diluting nuances between firing strengths.

As discussed earlier, \CuiEtAl~observed that \texttt{softmax} is present in the same NFNs used in this paper. If layer normalization were applied to the fuzzy logic rules' preliminary firing levels before Eq.~\ref{eq:norm_softmax}, then their firing levels may be elevated from near-zero (Figure~\ref{fig:rule_cod_layer_norm_softmax}). This has also helped maintain their relative activation, unlike Eq.~\ref{eq:mu_rule_mean} in Figure~\ref{fig:rule_cod_no_layer_norm_softmax}. The stark contrast between Eq.~\ref{eq:mu_rule_sum} and \CuiEtAl's Eq.~\ref{eq:mu_rule_mean} mostly disappears, but \CuiEtAl's Eq.~\ref{eq:mu_rule_mean} provides a small nudge in favor of activating the rules.

However, interesting behavior occurs when no layer normalization is applied, but 1.5-\texttt{entmax} is used instead of \texttt{softmax} (Figure~\ref{fig:rule_cod_no_layer_norm_entmax}). The activations calculated by \CuiEtAl's Eq.~\ref{eq:mu_rule_mean} followed by 1.5-\texttt{entmax} are no longer strictly uniform and reveal a bit more discernibility with respect to each other. On the contrary, rule activations according to the standard Eq.~\ref{eq:mu_rule_sum} followed by 1.5-\texttt{entmax} show a significant increase in their applicability.

If layer normalization and 1.5-\texttt{entmax} are both incorporated (Figure~\ref{fig:rule_cod_layer_norm_entmax}), no small set of fuzzy logic rules dominates like in Figure~\ref{fig:rule_cod_no_layer_norm_entmax}. The only difference is the incorporation of layer normalization before applying 1.5-\texttt{entmax}, which appears to have reduced the activations to nearly identical behavior as shown in Figure~\ref{fig:rule_cod_layer_norm_softmax}. This suggests the presence of both layer normalization with 1.5-\texttt{entmax} is not advantageous over simply using layer normalization with \texttt{softmax}, but this combination is still explored in our experimentation.

\footnotesize
\bibliography{references}






\end{document}